\documentclass[10pt,twocolumn,letterpaper]{article}

\usepackage{iccv}
\usepackage{times}
\usepackage{epsfig}
\usepackage{graphicx}
\usepackage{amsmath}
\usepackage{amssymb}
\usepackage{float}
\usepackage{caption}
\usepackage{array}
\usepackage{xcolor}
\usepackage{tablefootnote}
\usepackage{microtype}
\usepackage[T1]{fontenc}
\usepackage{chngcntr}
\usepackage[utf8]{inputenc}
\newcolumntype{C}[1]{>{\centering\arraybackslash}p{#1}}

\usepackage[pagebackref=true,breaklinks=true,letterpaper=true,colorlinks,bookmarks=false]{hyperref}

\iccvfinalcopy %

\def\iccvPaperID{2441} %

\ificcvfinal\fi

\begin{document}

\title{Erasing Concepts from Diffusion Models}

\author{Rohit Gandikota$^{*,1}$ \qquad Joanna Materzy\'nska$^{*,2}$ \qquad Jaden Fiotto-Kaufman$^{1}$ \quad David Bau$^{1}$ \vspace{3pt} \\ 
$^{1}$Northeastern University \qquad $^{2}$Massachusetts Institute of Technology\\
$^{1}${\tt\small [gandikota.ro, fiotto-kaufman.j, davidbau]@northeastern.edu} \quad $^{2}${\tt\small jomat@mit.edu}
}

\twocolumn[{%
\renewcommand\twocolumn[1][]{#1}%
\maketitle%
\vspace{-0.3in}%
\begin{center}
    \centering\includegraphics[width=\linewidth]{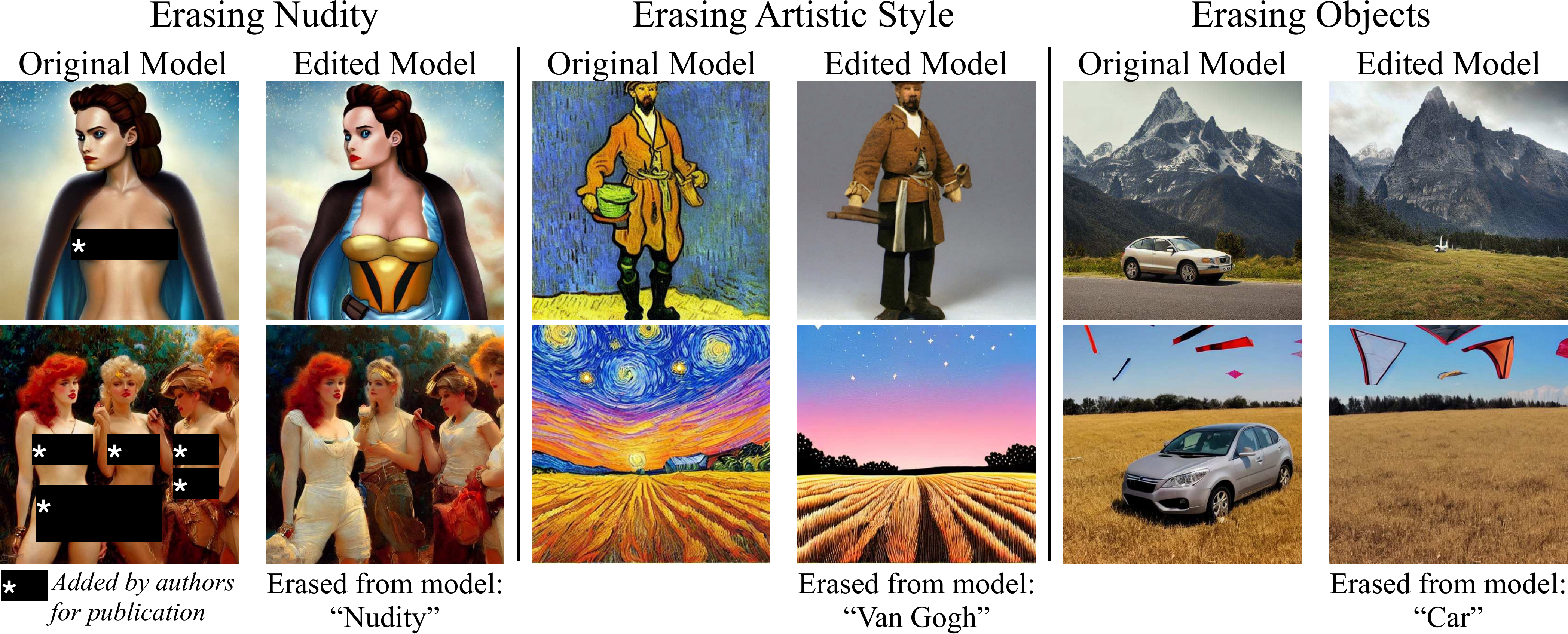}%
        \vspace{-0.05in}
        \captionof{figure}{Given only a short text description of an undesired visual concept and no additional data, our method fine-tunes model weights to erase the targeted concept. Our method can avoid NSFW content, stop imitation of a specific artist's style, or even erase a whole object class from model output, while preserving the model's behavior and capabilities on other topics.}%
    \label{fig:mainfig}
\end{center}
}]
\maketitle
\def\thefootnote{*}\footnotetext{Equal contribution}\def\thefootnote{\arabic{footnote}}

\ificcvfinal\thispagestyle{empty}\fi

\begin{abstract}
\vspace{-0.05in}
Motivated by concerns that large-scale diffusion models can produce undesirable output such as sexually explicit content or copyrighted artistic styles, we study erasure of specific concepts from diffusion model weights.  We propose a fine-tuning method that can erase a visual concept from a pre-trained diffusion model, given only the name of the style and using negative guidance as a teacher. We benchmark our method against previous approaches that remove sexually explicit content and demonstrate its effectiveness, performing on par with Safe Latent Diffusion and censored training.
To evaluate artistic style removal, we conduct experiments erasing five modern artists from the network and conduct a user study to assess the human perception of the removed styles. Unlike previous methods, our approach can remove concepts from a diffusion model permanently rather than modifying the output at the inference time, so it cannot be circumvented even if a user has access to model weights. Our code, data, and results are available at  \href{https://erasing.baulab.info}{erasing.baulab.info}.%
\vspace{-14pt} %
\end{abstract}%

\section{Introduction}
Recent text-to-image generative models have attracted attention due to their remarkable image quality and seemingly infinite generation capabilities. These models are trained on vast internet datasets, which enables them to imitate a wide range of concepts. However, some concepts learned by the model are undesirable, including copyrighted content and pornography, which we aim to avoid in the model's output~\cite{dalle2022modelcard,imagen2022webpage,rando2022red}. In this paper, we propose an approach for selectively removing a single concept from a text-conditional model's weights after pretraining. Prior approaches have focused on dataset filtering~\cite{rombach2022sd20}, post-generation filtering \cite{rando2022red}, or inference guiding \cite{schramowski2022safe}. Unlike data filtering methods, our method does not require retraining, which is prohibitive for large models. Inference-based methods can censor~\cite{rando2022red} or steer the output away from undesired concepts effectively~\cite{schramowski2022safe}, but they can be easily circumvented. In contrast, our approach directly removes the concept from the model's parameters, making it safe to distribute its weights.

The open-source release of the Stable Diffusion text-to-image diffusion model has made image generation technology accessible to a broad audience.  To limit the generation of unsafe images, the first version was bundled with a simple NSFW filter to censor images if the filter is triggered~\cite{rando2022red}, yet since both the code and model weights are publicly available, it is easy to disable the filter~\cite{smith2022howto}. In an effort to prevent the generation of sensitive content, the subsequent SD 2.0 model is trained on data filtered to remove explicit images, an experiment consuming 150,000 GPU-hours of computation~\cite{sd142022modelcard}  over the 5-billion-image LAION dataset~\cite{schuhmann2022laion}. The high cost of the process makes it challenging to establish a causal connection between specific changes in the data and the capabilities that emerge, but users report that removing explicit images and other subjects from the training data may have had a negative impact on the output quality~\cite{rombach2022sd20}. And despite the effort, explicit content remains prevalent in the model's output: when we evaluate generation of images using prompts from the 4,703 prompts of the Inappropriate Image Prompts (I2P) benchmark~\cite{schramowski2022safe}, we find that the popular SD 1.4 model produces 796 images with exposed body parts identified by a nudity detector, while the new training-set-restricted SD 2.0 model produces 417 (Figure~\ref{fig:nudity_barplot}). %

Another major concern regarding the text-to-image models is their ability to imitate potentially copyrighted content. Not only is the quality of the AI-generated art on par with the human-generated art \cite{roose22ai}, it can also faithfully replicate an artistic style of real artists. Users of Stable Diffusion~\cite{rombach2022high} and other large-scale text-to-image synthesis systems have discovered that prompts such as ``art in the style of [\emph{artist}]'' can mimic styles of specific artists, potentially devaluing original work. Copyright concerns of several artists has led to a lawsuit against the makers of Stable Diffusion~\cite{andersen2023stability}, raising new legal issues~\cite{setty2023suit}; the courts have yet to rule on these cases. Recent work~\cite{shan2023glaze} aims to protect the artist by applying an adversarial perturbation to artwork before posting it online to prevent the model from imitating it. That approach, however, cannot remove a learned artistic style from a pretrained model.

In response to safety and copyright infringement concerns, we propose a method for erasing a concept from a text-to-image model. %
Our method, Erased Stable Diffusion (ESD), fine-tunes the model's parameters using only undesired concept descriptions and no additional training data. Unlike training-set censorship approaches, our method is fast and does not require training the whole system from scratch. Furthermore, our method can be applied to existing models without the need to modify input images \cite{shan2023glaze}. Unlike the post-filtering \cite{rando2022red} or simple blacklisting methods, erasure cannot be easily circumvented, even by users who have access to the parameters.
We benchmark our method on removing offensive content and find that it is as effective as Safe Latent Diffusion \cite{schramowski2022safe} for removing offensive images. We also test the ability of our method to remove an artistic style from the model. We conduct a user study to test the impact of erasure on user perception of the remove artist's style in output images, as well as the interference with other artistic styles and their impact on image quality.  Finally, we also test our method on erasure of complete object classes.

\section{Related Works}
\textbf{Undesirable image removal.} Previous work to avoid undesirable image output in generative models has taken two main approaches: The first is to censor images from the training set, for example, by removing all people~\cite{nichol2021glide}, or by more narrowly curating data to exclude undesirable classes of images~\cite{schuhmann2022laion,dalle2022modelcard,sdv22022modelcard}.  Dataset removal has the disadvantage that the resources required to retrain large models makes it a very costly way to respond to problems discovered after training; also large-scale censorship can produce unintended effects~\cite{oconnor2022stable}. The second approach is post-hoc, modifying output after training using classifiers~\cite{bedapudi2022nudenet,laborde2022nsfw,rando2022red}, or by adding guidance to the inference process~\cite{schramowski2022safe}; such methods are efficient to test and deploy, but they are easily circumvented by a user with access to parameters~\cite{smith2022howto}.  We compare both previous approaches including Stable Diffusion 2.0~\cite{rombach2022sd20}, which is a complete retraining of the model on a censored training set, and Safe Latent Diffusion~\cite{schramowski2022safe}, which the state-of-the-art guidance-based approach.  The focus of our current work is to introduce a third approach: we tune the model parameters using a guidance-based model-editing method, which is both fast to employ and also difficult to circumvent.

\textbf{Image cloaking.}  Another approach to protecting images from imitation by large models is for an artist to cloak images by adding adversarial perturbations before posting them on the internet.  Cloaking allows artists to effectively hide their work from a machine-learned model during training or inference by adding perturbations that cause the model to confuse the cloaked image with an unrelated image~\cite{salman2023raising} or an image with a different artistic style~\cite{shan2023glaze}; the method is a promising way for an artist to self-censor their own content from AI training sets while still making their work visible to humans.
Our paper addresses a different problem than the problem addressed by cloaking: we ask how a model creator can erase an undesired visual concept without active self-censorship by content providers.
\begin{figure*}
    \centering
    \includegraphics[width=1\linewidth]{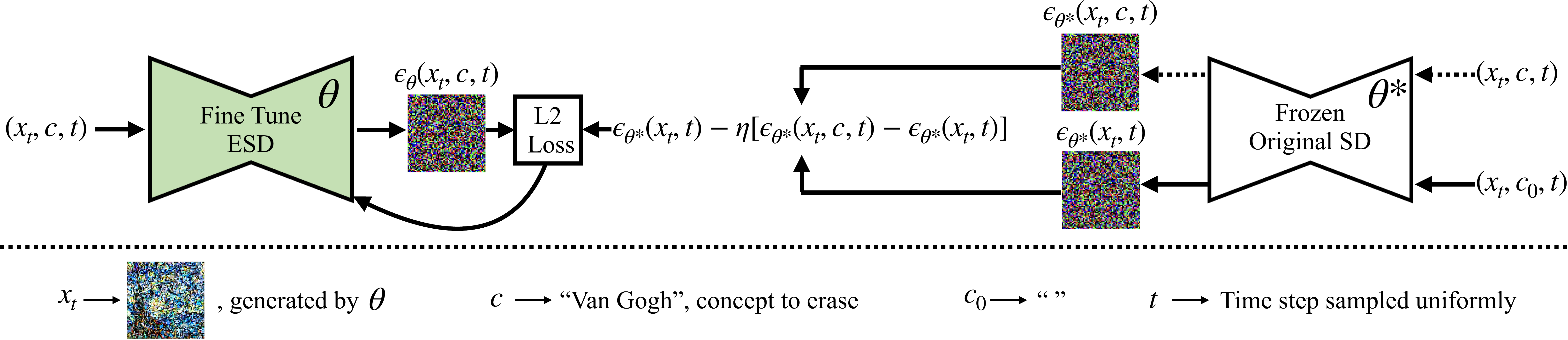}
    \caption{
    The optimization process for erasing undesired visual concepts from pre-trained diffusion model weights involves using a short text description of the concept as guidance. 
    The ESD model is fine-tuned with the conditioned and unconditioned scores obtained from frozen SD model to guide the output away from the concept being erased. The model learns from its own knowledge to steer the diffusion process away from the undesired concept. }
    \label{fig:architecture}
\end{figure*}

\textbf{Model editing.}  As the cost of training grows, there has been increasing interest in lightweight model-editing methods that alter the behavior of large-scale generative models given little or no new training data.  In text generators, a model's knowledge of facts can be edited based on a single statement of the fact by modifying specific neurons~\cite{dai2022knowledge} or layers~\cite{meng2022locating}, or by using hypernetworks~\cite{de2021editing,mitchell2021fast}.  In image synthesis, a generative adversarial network (GAN) can be edited using a handful of words~\cite{gal2022stylegan}, a few sketches~\cite{wang2021sketch}, warping gestures~\cite{wang2022rewriting}, or copy-and-paste~\cite{bau2020rewriting}.  Recently, it has been shown that text-conditional diffusion models can be edited by associating a token for a new subject trained using only a handful of images~\cite{gal2023an, kumari2022customdiffusion, ruiz2022dreambooth}. Unlike previous methods that add or modify the appearance of objects, the goal of our current work is to erase a targeted visual concept from a diffusion model given only a single textual description of the concept, object, or style to be removed.

\textbf{Memorization and unlearning.} While the traditional goal of machine learning is to generalize without memorization, large models are capable of exact memorization if specifically trained to do so~\cite{zhang2021understanding}, and unintentional memorization has also been observed in large-scale settings~\cite{carlini2019secret,carlini2023quantifying}, including diffusion models~\cite{somepalli2022diffusion}.  The possibility of such exact memorization has driven privacy and copyright concerns and has led to work in machine unlearning~\cite{sekhari2021remember,bourtoule2021machine,golatkar2020eternal}, which aims to modify a model to behave as if particular training data had not been present. However, these methods are based on the assumption that the undesired knowledge corresponds to an identifiable set of training data points.  The problem we tackle in this paper is very different from the problem of unlearning specific training data because rather than simulating the removal of a known training item, our goal is to erase a high-level visual concept that may have been learned from a large and unknown subset of the training data, such as the appearance of nudity, or the imitation of an artist's style.

\textbf{Energy-based composition.}  Our work is inspired by the observation~\cite{du2020compositional,du2021unsupervised} that set-like composition can be performed naturally on energy-based models and diffusion counterparts~\cite{liu2022compositional} naturally via arithmetic on the score or the noise predictions.  Score-based composition is also the basis for classifier-free-guidance~\cite{ho2022classifier}.  Like previous works, we treat ``A and not B'' as the difference between log probability densities for A and B; a similar observation  has been used to reduce the undesirable output of both language models~\cite{schick2021self} and vision generators~\cite{schramowski2022safe}.  Unlike previous work that applies composition at inference time, we introduce the use of score composition as a source of unsupervised training data to teach a fine-tuned model to erase an undesired concept from model weights. %

\section{Background}
\subsection{Denoising Diffusion Models}
Diffusion models are a class of generative models that learn the distribution space as a gradual denoising process~\cite{sohl2015diffusion,ho2020denoising}. Starting from sampled Gaussian noise, the model gradually denoises for $T$ time steps until a final image is formed. In practice, the diffusion model predicts noise $\epsilon_t$ at each time step $t$ that is used to generate the intermediate denoised image $x_t$; where $x_T$ corresponds to the initial noise and $x_0$ corresponds to the final image. This denoising process is modeled as a Markov transition probability.
\begin{align}
    p_{\theta}(x_{T:0}) = p(x_T)\prod_{t=T}^{1}p_{\theta}(x_{t-1} | x_t)
\end{align}

\subsection{Latent Diffusion Models}
Latent diffusion models (LDM)~\cite{rombach2022high} improve efficiency by operating in a lower dimensional latent space $z$ of a pretrained variational autoencoder with encoder $\mathcal{E}$ and decoder $\mathcal{D}$. During training, for an image $x$, noise is added to its encoded latent, $z = \mathcal{E}(x)$ leading to $z_t$ where the noise level increases with $t$. LDM process can be interpreted as a sequence of denoising models with identical parameters $\theta$ that learn to predict the noise $\epsilon_\theta(z_t, c, t)$ added to $z_t$ conditioned on the timestep $t$ as well as a text condition $c$. The following objective function is optimized:
\begin{align}
    \mathcal{L} = \mathbb{E}_{z_t\in\mathcal{E}(x)\textit{,} t\textit{,} c \textit{,} \epsilon\sim\mathcal{N}(0,1)} [\| \epsilon - \epsilon_\theta(z_t, c, t)\|_2^2]
\end{align}

Classifier-free guidance is a technique employed to regulate image generation, as described in Ho et al. \cite{ho2022classifier}. This method involves redirecting the probability distribution towards data that is highly probable according to an implicit classifier $p(c | z_t)$. This approach is used during inference and requires that the model be jointly trained on both conditional and unconditional denoising. The conditional and unconditional scores are both obtained from the model during inference. The final score $\Tilde{\epsilon}_\theta(z_t, c, t)$ is then directed towards the conditioned score and away from the unconditioned score by utilizing a guidance scale $\alpha > 1$.
\begin{align}
    \Tilde{\epsilon}_\theta(z_t, c, t) = \epsilon_\theta(z_t,t) + \alpha(\epsilon_\theta(z_t, c, t) - \epsilon_\theta(z_t,t))
\end{align}

The inference process starts from a Gaussian noise $z_T \sim \mathcal{N}(0,1)$ and is denoised with the $\Tilde{\epsilon}_\theta(z_T, c, T)$ to get $z_{T-1}$. This process is done sequentially till $z_0$ and is transformed to image space using the decoder $x_0 \gets \mathcal{D}(z_0)$.

\section{Method}

The goal of our method is to erase concepts from text-to-image diffusion models using its own knowledge and no additional data. Therefore, we consider fine-tuning a pre-trained model rather than training a model from scratch. 
We focus on Stable Diffusion (SD) \cite{rombach2022high}, an LDM that consists of 3 subnetworks: a text encoder $\mathcal{T}$, a diffusion model (U-Net) $\theta^*$ and a decoder model $\mathcal{D}$. We shall train new parameters $\theta$.

Our approach involves editing the pre-trained diffusion U-Net model weights to remove a specific style or concept. We aim to reduce the probability of generating an image x according to the likelihood that is described by the concept, scaled by a power factor $\eta$. 
\begin{align}
    P_{\theta}(x) \propto \frac{P_{\theta^*}(x)}{P_{\theta^*}(c|x)^\eta}
\end{align}
Where $P_{\theta^*}(x)$ represents the distribution generated by the original model and $c$ represents the concept to erase. Expanding $P(c|x)=\frac{P(x|c)P(c)}{P(x)}$, the gradient of the log probability $\nabla \log P_{\theta}(x)$ would be proportional to:
\begin{align}
     \nabla \log P_{\theta^*}(x) - \eta(\nabla \log P_{\theta^*}(x|c) - \nabla \log P_{\theta^*}(x))
     \label{eq:newscore}
\end{align}

Based on Tweedie's formula \cite{efron2011tweedie} and the reparametrization trick of \cite{ho2020denoising},  we can introduce a time-varying noising process and express each score (gradient of log probability) as a denoising prediction $\epsilon(x_t,c,t)$. Thus Eq.~\ref{eq:newscore} becomes:
\begin{align}
   \epsilon_\theta(x_t, c, t) \gets \epsilon_{\theta^*}(x_t, t) -\eta[\epsilon_{\theta^*}(x_t, c, t) - \epsilon_{\theta^*}(x_t, t)]
   \label{eq:objective}
\end{align}
This modified score function moves the data distribution to minimize the generation probability of images $x$ that can be labeled as $c$. The objective function in Equation~\ref{eq:objective} fine-tunes the parameters $\theta$ such that $\epsilon_\theta(x_t, c, t)$ mimics the negatively guided noise. That way, after the fine-tuning, the edited model's conditional prediction is guided away from the erased concept.

Figure~\ref{fig:architecture} illustrates our training process. %
We exploit the model's knowledge of the concept to synthesize training samples, thereby eliminating the need for data collection. Training uses several instances of the diffusion model, with one set of parameters frozen ($\theta^*$)  while training the other set of parameters ($\theta$) to erase the concept. We sample partially denoised images $x_t$ conditioned on $c$ using $\theta$, then we perform inference on the frozen model $\theta^*$ twice to predict the noise, once conditioned on $c$ and the other unconditioned. Finally, we combine these two predictions linearly to negate the predicted noise associated with the concept, and we tune the new model towards that new objective.

\subsection{Importance of Parameter Choice}
\label{intuition_section}

\begin{figure}
  \centering
  \includegraphics[width=.97\linewidth]{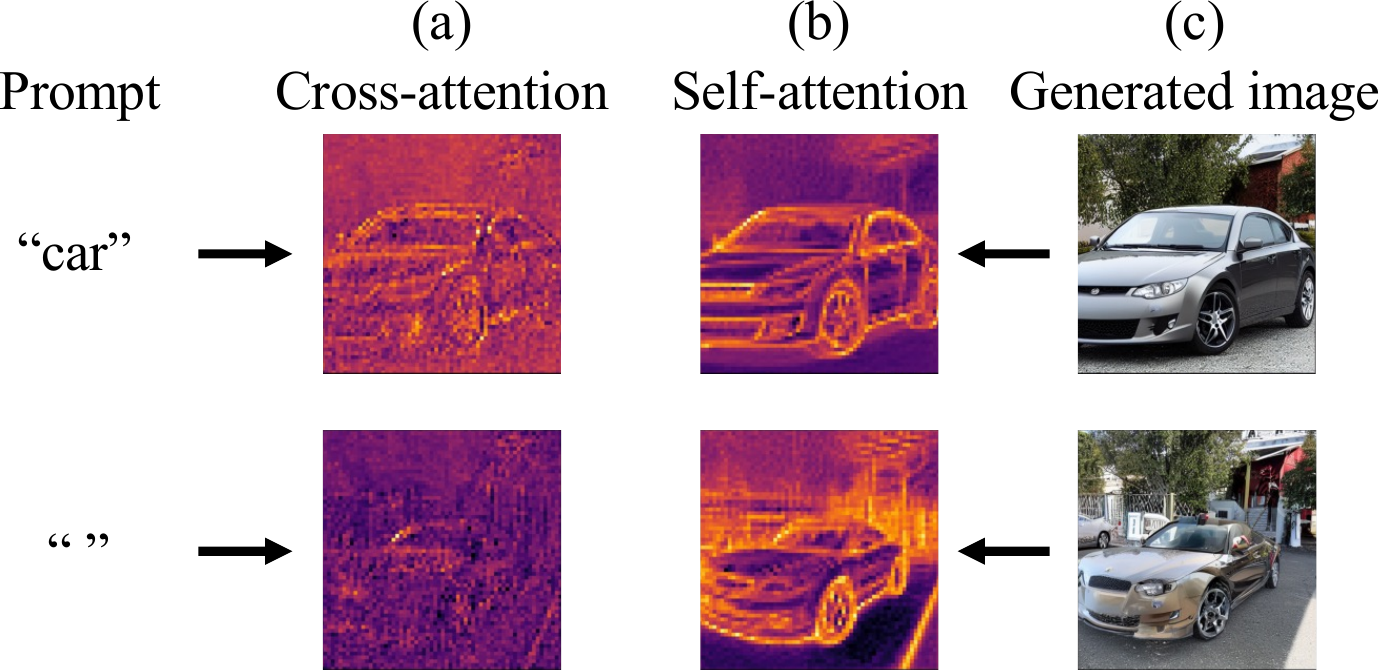}
  \caption{When comparing generation of two similar car images conditioned on different prompts, self-attention (b) contributes to the features of a car regardless of the presence of the word ``car'' in the prompt, while the contribution of cross-attention (a) is linked to the presence of the word. Heatmaps show local contributions of the first attention modules of the 3rd upsampling block of the Stable Diffusion U-net while generating the images (c).}
  \label{fig:attentionmap}
\end{figure}

The effect of applying the erasure objective (\ref{eq:objective}) depends on the subset of parameters that is fine-tuned.  %
The main distinction is between cross-attention parameters and non-cross-attention parameters.  Cross-attention parameters, illustrated in Figure~\ref{fig:attentionmap}a,  serve as a gateway to the prompt, directly depending on the text of the prompt, while other parameters (Figure~\ref{fig:attentionmap}b) tend to contribute to a visual concept even if the concept is not mentioned in the prompt.

Therefore we propose fine tuning the cross attentions, ESD-x, when the erasure is required to be controlled and specific to the prompt, such as when a named artistic style should be erased. Further, we propose fine tuning unconditional layers (non-cross-attention modules), ESD-u, when the erasure is required to be independent of the text in the prompt, such as when the global concept of NSFW nudity should be erased.  We refer to cross-attention-only fine-tuning as ESD-x-$\eta$ (where $\eta$ refers to the strength of the negative guidance), and we refer to the configuration that tunes only non-cross-attention parameters as ESD-u-$\eta$. For simplicity, we write ESD-x and ESD-u when $\eta=1$.

\begin{figure}
  \centering
  \includegraphics[width=1\linewidth]{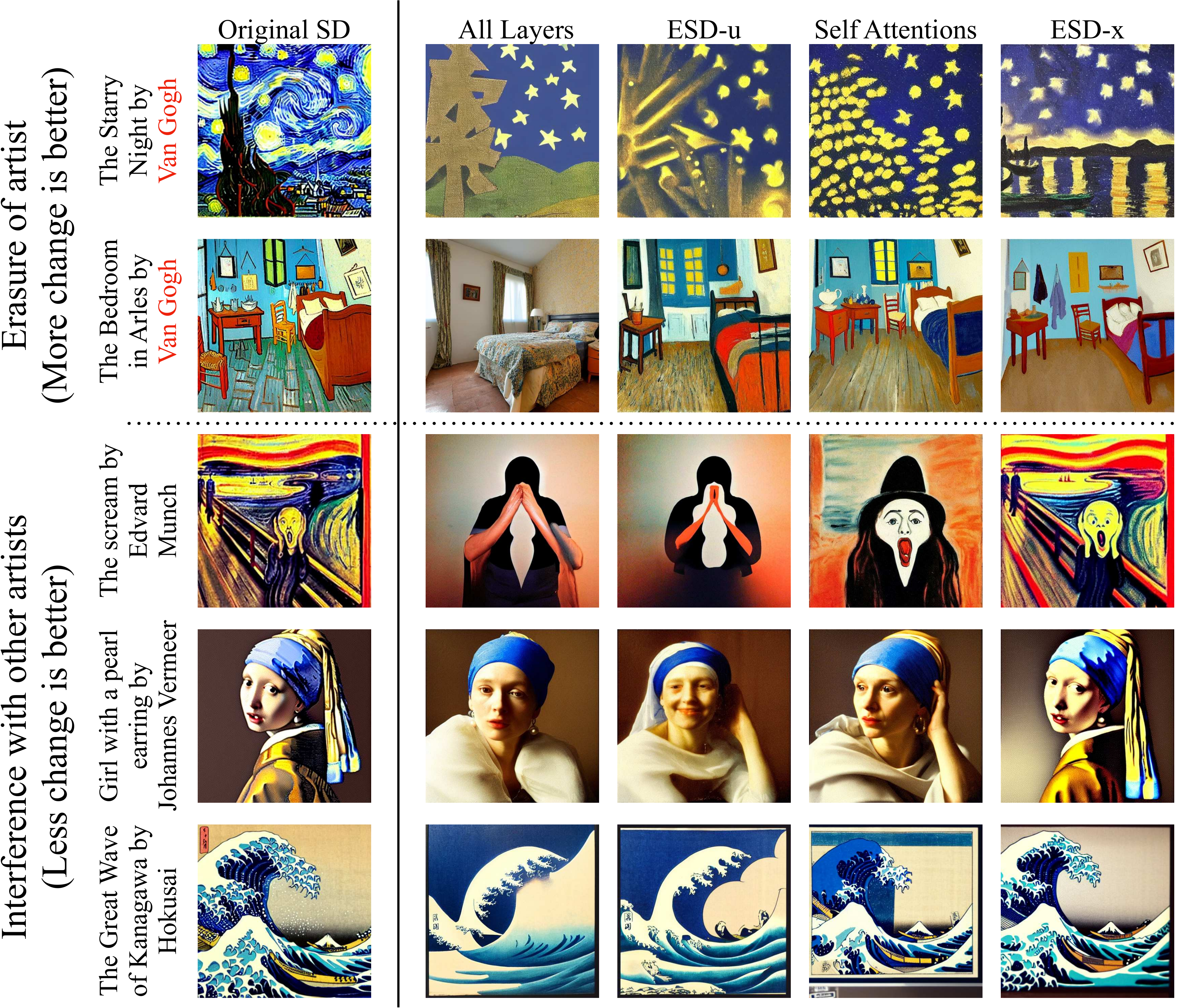}
  \caption{Modifying the cross-attention weights, ESD-x, shows negligible interference with other styles (bottom 3 rows) and is thus well-suited for erasing art styles. In contrast, altering the non-cross-attention weights, ESD-u, has a global erasure effect (all rows) on the visual concept and is better suited for removing nudity or objects.}
   \label{fig:layerablation}
\end{figure}

The effects of parameter choices on artist style removal are illustrated in  Figure~\ref{fig:layerablation}: when erasing the ``Van Gogh'' style ESD-u and other unconditioned parameter choices erase aspects of the style globally, erasing aspects of Van Gogh's style from many artistic styles other than Van Gogh's. On the other hand, tuning the the cross-attention parameters only (ESD-x) erases the distinctive style of Van Gogh specifically when his name is mentioned in the prompt, keeping the interference with other artistic styles to a minimum.

Conversely, when removing NSFW content it is important that the visual concept of ``nudity'' is removed globally, \emph{especially} in cases when nudity is \emph{not} mentioned in the prompt. To measure those effects we evaluate on a data set that include many prompts that do not explicitly mention NSFW terms (Section~\ref{sec:nsfw-experiment}).  We find that ESD-u performs best in this application; full quantitative ablations over different parameter sets  are included in Appendix~\ref{appendix:nudity_application}.

\section{Experiments}
We train all our models for 1000 gradient update steps on a batch size of 1 with learning rate 1e-5 using the Adam optimizer. Depending on the concept we want to remove (\ref{intuition_section}), the ESD-x method fine-tunes the cross-attention and ESD-u fine tunes the unconditional weights of the U-Net module in Stable Diffusion (our experiments use version 1.4 unless specified otherwise).  Baseline methods are:
\begin{itemize}
\itemsep-0.35em 
    \item SD (pretrained Stable Diffusion),
    \item SLD (Safe Latent Diffusion) \cite{schramowski2022safe}, to adapt the method to our experiment, we substitute the concept we want to erase from the model for the original safety concepts.
    \item SD-Neg-Prompt (Stable Diffusion with Negative Prompts), an inference technique in the community, that aims to steer away from unwanted effects in an image. We adapt this method by using the artist's name as the negative prompt.
\end{itemize}

\subsection{Artistic Style Removal}
\begin{figure*}
  \centering
  \includegraphics[width=\linewidth]{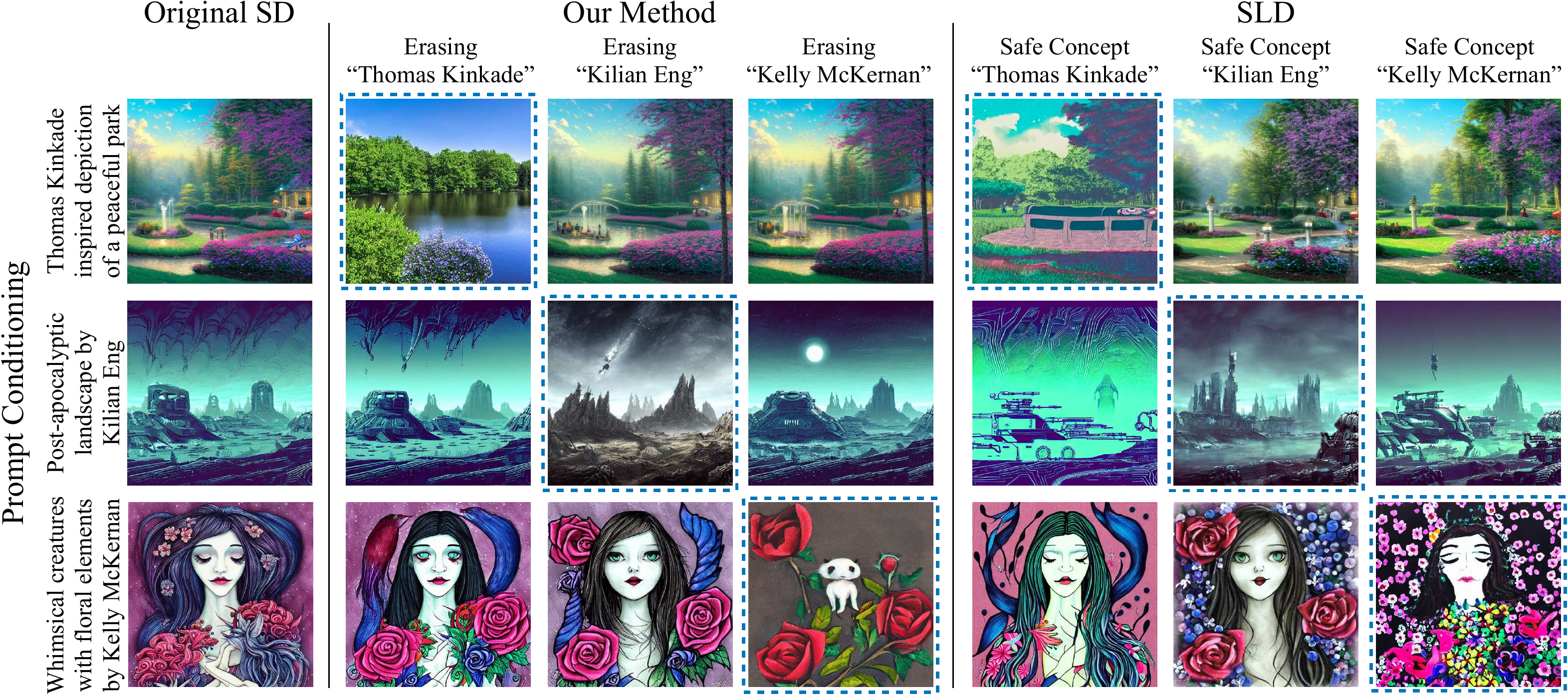}
   \caption{Our method has a better erasure on intended style with a minimal interference compared to SLD \cite{schramowski2022safe}. The images enclosed in blue dotted borders are the intended erasure, and the off-diagonal images show effect on untargeted styles.}
   \label{fig:artistsintereference}
\end{figure*}

\subsubsection{Experiment Setup}
\label{artist-imitation-study}
To analyze imitation of art among contemporary practicing artists, we consider 5 modern artists and artistic topics; Kelly McKernan, Thomas Kinkade, Tyler Edlin, Kilian Eng and the series ``Ajin: Demi-Human,'' which have been reported to be imitated by Stable Diffusion. While we did not observe the model making direct copies of specific original artwork, it is undeniable that these artistic styles have been captured by the model. To study this effect, we demonstrate qualitative results in Fig \ref{fig:artistsintereference} and conduct a user study to measure the human perception on the artistic removal effect. 
Our experiments validate the observation that the particular artist-specific style is removed from the model, while the content and structure of the prompt is preserved (Fig \ref{fig:artistsintereference}) with minimal interference on other artistic styles. For more image examples, please refer to Appendix~\ref{sec:exp}.

\label{style-removal}
\begin{figure}
  \centering
  \includegraphics[width=1\linewidth]{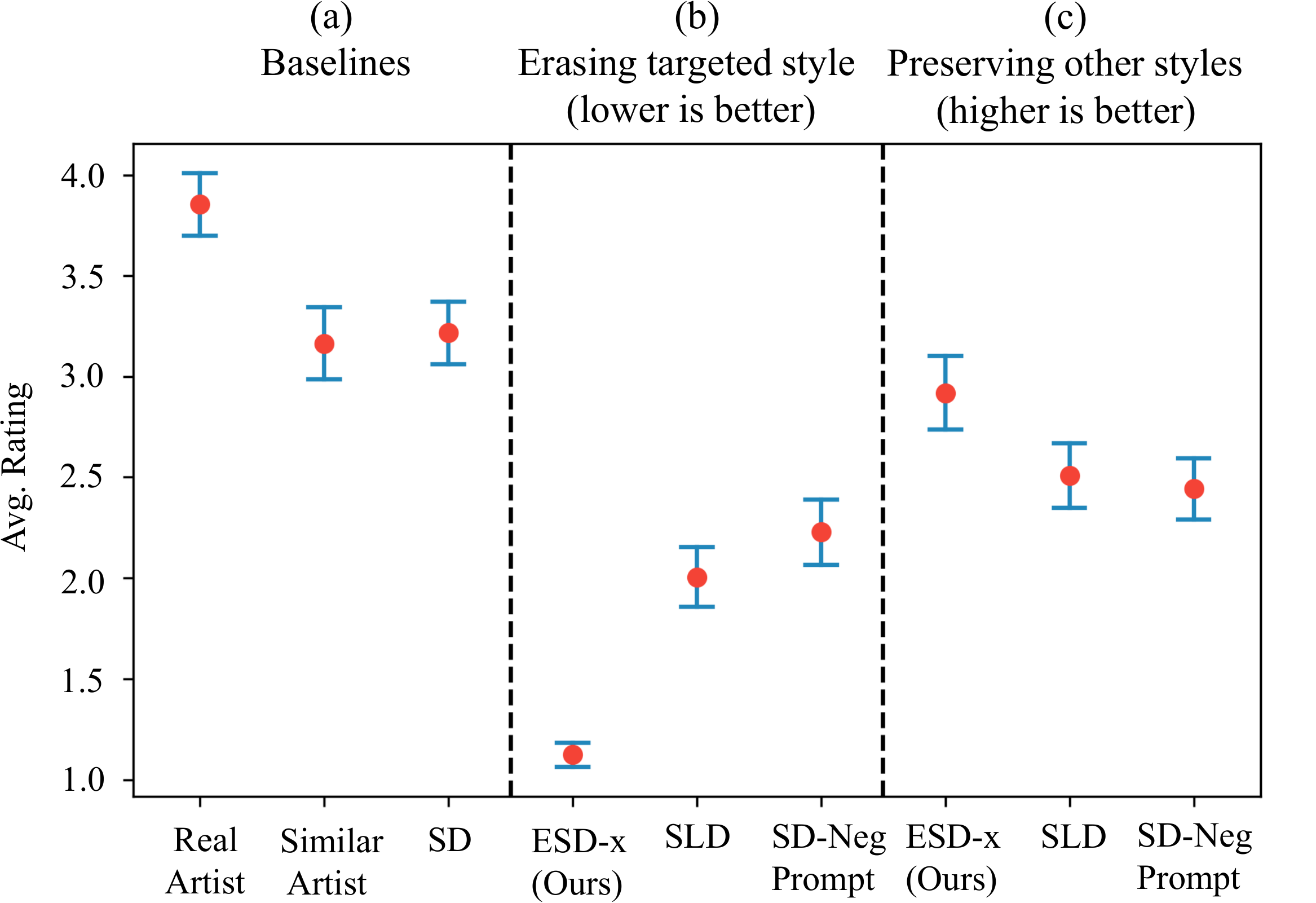}
   \caption{User study ratings (with ± 95\% confidence intervals) show that our method erases the intended style better than the baselines. The rating (1-5) represent the similarity of the images compared to original artist style (5 being most similar). With higher ratings for images from similar style artists, the study shows that style is highly subjective. }
   \label{fig:artistsconfusion}
   \vspace{-1em}
\end{figure}
\subsubsection{Artistic Style Removal User Study}
\label{sec:art-user-study}
To measure the human perception of the effectiveness of the removed style, we conducted a user study.
For each artist, we collect 40 images of art created by those artists, using Google Image Search to identify top-ranked works. Then for each artist, we also compose 40 generic text prompts that invoke the artist's style, and we use Stable Diffusion to generate images for each artist using these prompts, for example: “Art by
[\textit{artist}]”, “A design of [\textit{artist}]”, “An image in the style of [\textit{artist}]”, “A 
reproduction of the famous art of [\textit{artist}]”. We also evaluate images from edited diffusion models, as described in Section~\ref{style-removal} as well as the baseline models. The images were generated with 4 seeds per prompt (same seeds were used for all methods) resulting in a dataset of 1000 images. Furthermore, we also included images from a similar human artist for each of the five artists. We pair real artist with similar real artist as follows: (Kelly McKernan, Kirbi Fagan), (Thomas Kinkade, Nicky Boehme), (Ajin: Demi Human, Tokyo Ghoul), (Tyler Edlin, Feng Zhu), (Kilian Eng, Jean Giraud). For each similar artist, we collected 12-25 works to use in our study. 

\begin{figure*}
  \centering
  \includegraphics[width=1\linewidth]{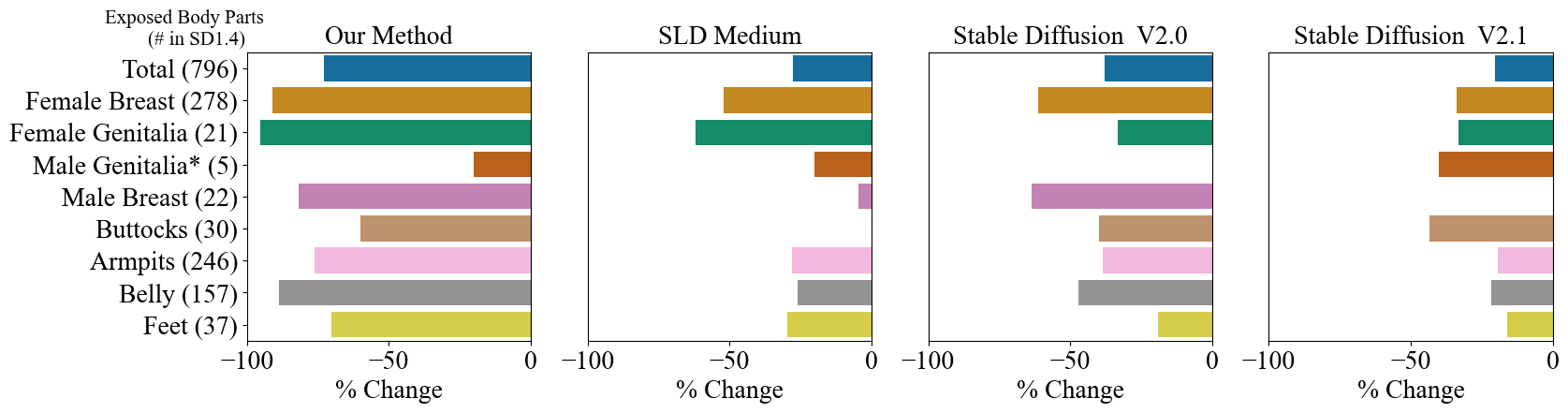}
   \caption{Our method effectively removes nudity content from Stable Diffusion on I2P data, outperforming inference method, SLD \cite{schramowski2022safe} and model trained on NSFW filtered dataset, SD-V2.0 \cite{rombach2022sd20}. (SD-V2.1, also shown, filters less aggressively.) The figure shows the percentage change in the nudity-classified samples compared to the original SD-V1.4 model. The SD v1.4 produces 796 images with exposed body parts on the test prompts, and our method reduces this total to 134. }
   \label{fig:nudity_barplot}
\end{figure*}
In our study, participants were presented with a set of five real artwork images along with an additional image. The additional image was either a real artwork from the same artist or a similar artist, or a synthetic image generated using a prompt involving the artist name with our method (ESD-x) or other baseline methods (SLD and SD-Neg-Prompt) applied to remove the artist  or a random different artist. Participants were asked to estimate, on a five-point Likert scale, their confidence level that the experimental image was created by the same artist as the five real artworks.

Our study involved 13 total participants, with an average of 170 responses per participant. We evaluated the effectiveness of the ESD-x method for removing the style of five modern artists and measuring the resemblance of generated artistic images to real images. Additionally, we assessed the amount of interference introduced by our method in comparison to other baseline methods, measuring the extent to which other artistic styles were affected.

The findings of our investigation are presented in Figure \ref{fig:artistsconfusion}. Interestingly, even for the genuine artwork, there is some level of uncertainty about its authenticity. The average rating for the original images is 3.85, the average rating for artists similar to the selected artists is 3.16, while the average rating for the AI-generated art is 3.21, indicating that the AI-duplicates are rated higher than similar genuine artwork. These outcomes reinforce our observations that the model effectively captures the style of an artist. All three removal techniques effectively decrease the perceived artistic style, with average ratings of 1.12, 2.00, and 2.22 for ESD-x, SLD and SD-Neg-Prompt respectively.

In addition to assessing the removal of an artistic style, we are also interested in evaluating the interference of our method with different artistic styles. To accomplish this, we present our research participants with images produced using a text prompt referring to an artist that was not erased in a model generated with an erased artist. We compare our method to SLD and SD-Neg-Prompt in this experiment. The outcomes, presented in Figure \ref{fig:artistsintereference}, indicate that users are most likely to consider images generated using our method to be genuine artwork, as opposed to those generated using other removal techniques, indicating that our method does not interfere with other artistic styles as a result of removing an artist. It is worth noting that, unlike the two baselines, our method modifies the model permanently, rather than being an inference approach.

Our method can also be applied to erase single works of art rather than entire artistic styles; we describe and analyze this variant in Appendix~\ref{sec:imageESD}.

\subsection{Explicit Content Removal}
\label{sec:nsfw-experiment}

\begin{table}
    \centering
    \begin{tabular}{lccc}
        \textbf{ Method} & \textbf{FID-30k} & \textbf{CLIP} & \tabularnewline
         \hline
         REAL & - & 0.1561  \tabularnewline
         SD  & 14.50 & 0.1592  \tabularnewline
         SLD-Medium\tablefootnote{These numbers are taken from the SLD paper \cite{schramowski2022safe}. In our experiments, we find the FIDs to be 18.71 and 25.29 for Medium and Max respectively} &  16.90  & 0.1594 &\tabularnewline %
         SLD-Max & 18.76 & 0.1583 \tabularnewline %
         ESD-u & 13.68 & 0.1585 \tabularnewline
         ESD-u-3 & 17.27 & 0.1586 \tabularnewline
    \end{tabular}
    \caption{Our method shows better image fidelity performance compared to SLD and Stable Diffusion on COCO 30k images generated with inference guidance $\alpha = 7.5$. All the methods show good CLIP score consistency with SD.  }
    \label{tab:fid}
\end{table}

Recent works have addressed the challenge of NSFW content restriction either through inference modification \cite{schramowski2022safe}, post-production classification based restriction \cite{rombach2022high} or re-training the entire model with NSFW restricted subset of LAION dataset \cite{rombach2022sd20}. Inference and post-production classification based methods can be easily circumvented when the models are open-sourced \cite{smith2022howto}. Retraining the models on filtered data can be very expensive, and we find that such models (Stable Diffusion V2.0) are still capable of generating nudity, as seen in Figure~\ref{fig:nudity_barplot}.

Since the erasure of unsafe content like nudity requires the effect to be global and independent of text embeddings, we use ESD-u to erase "nudity". In Figure~\ref{fig:nudity_barplot}, we compare the percentage change in nudity classified samples with respect to Stable Diffusion v1.4. We study the effectiveness of our method with both inference method (SLD \cite{schramowski2022safe}) and filtered re-training methods (SD V2.0 \cite{rombach2022sd20}). For all the models, 4703 images are generated using I2P prompts from \cite{schramowski2022safe}. The images are classified into various nudity classes using the Nudenet~\cite{NudeNet2019Git} detector. For this analysis, we show results for our weak erasure scale of $\eta = 1$. We find that across all the classes, our method has a more significant effect in erasing nudity\footnote{We note that Nudenet \cite{NudeNet2019Git} has a higher false positive rate for male genitalia class. With manual cross verification, we find that both our method and SLD completely erases male genitalia, while SD V2.0 erases by 60\% }. For a more similar comparison study that was done by \cite{schramowski2022safe}, please refer to Appendix~\ref{appendix:nudity_application}.

To ensure that the erased model is still effective in generating safe content, we compare all the methods' performance on COCO 30K dataset prompts. We measure the image fidelity to show the quality and CLIP score to show specificity of the model to generate conditional images in table~\ref{tab:fid}. ESD-u refers to soft erasure with $\eta = 1$ and ESD-u-3 refers to stronger erasure with $\eta = 3$. Since COCO is a well curated dataset without nudity, this could be a reason for our method's better FID compared to SD. All the methods have similar CLIP scores as SD showing minimal effect specificity. \par

\subsection{Object Removal}
\label{sec:obj-removal}
In this section, we investigate the extent to which  method can also be used to erase entire object classes from the model. We prepare ten ESD-u models, each removing one class name from a subset of the ImageNet classes \cite{deng2009imagenet} (we investgiate the Imagenette \cite{howard2020fastai} subset which comprises ten easily identifiable classes). To measure the effect of removing both the targeted and untargeted classes, we generate 500 images of each class using base Stable Diffusion as well as each of the ten fine-tuned models using the prompt “an image of a  [\textit{class name}]”; then we evaluate the results by examining the top-1 predictions of a pretrained Resnet-50 Imagenet classifier.  Table \ref{tab:imagenette} displays quantitative results, comparing classification accuracy of the erased class in both the original Stable Diffusion model and our ESD-u model trained to eliminate the class. The table also shows the classification accuracy when generating the remaining nine classes. It is evident that our approach effectively removes the targeted classes in most cases, although there are some classes such as ``church'' that are more difficult to remove. Accuracy of untargeted classes remains high, but there is some interference, for example, removing ``French horn'' adds noticible distortions to other classes. Images showing the visual effects of object erasure are included in Appendix~\ref{appendix:object_application}.

\begin{table}
    \centering
\resizebox{\columnwidth}{!}{%
\footnotesize
\begin{tabular}{@{\extracolsep{4pt}}lcccc@{}}%
Class name & \multicolumn{2}{C{2cm}}{Accuracy of erased class} & \multicolumn{2}{C{2cm}}{Accuracy of other classes}
\tabularnewline
\cline{2-3}\cline{4-5} 
 & SD & ESD-u & SD & ESD-u\tabularnewline
\hline 
cassette player & 15.6 & 0.60 & 85.1 & 64.5\tabularnewline
chain saw & 66.0 & 6.0 & 79.6 & 68.2\tabularnewline
church & 73.8 & 54.2 & 78.7 & 71.6\tabularnewline
gas pump & 75.4 & 8.6 & 78.5 & 66.5\tabularnewline
tench & 78.4 & 9.6 & 78.2 & 66.6\tabularnewline
garbage truck & 85.4 & 10.4 & 77.4 & 51.5\tabularnewline
English springer & 92.5 & 6.2 & 76.6 & 62.6\tabularnewline
golf ball & 97.4 & 5.8 & 76.1 & 65.6\tabularnewline
parachute & 98.0 & 23.8 & 76.0 & 65.4\tabularnewline
French horn & 99.6 & 0.4 & 75.8 & 49.4\tabularnewline
\hline 
Average & 78.2 & 12.6 & 78.2 & 63.2
\end{tabular}
}
    \caption{Our method can cleanly erase many object concepts from a model, evidenced here a significant drop in classification accuracy of the concept while keeping the other class scores high. We measure the extent to which erasing an object class from the model affects the scores of other classes}
    \label{tab:imagenette}
\end{table}

\subsection{Effect of $\eta$ on Interference}
To measure the effect of $\eta$ on interference, we test three different "nudity"
 erased ESD-u-$\eta$ models' performance on 1000 images per each of the 10 Imagenette \cite{howard2020fastai} classes. Setting $\eta=10$ erases 92\% of nudity cases but reduces 1000-way classification accuracy by 34\% on object images, while $\eta=3$ erases 88\% and impacts objects by 14\%, and $\eta=1$ erases 83\% and impacts objects by 7\%. These results suggest that reducing the value of $\eta$ can mitigate interference, although reducing $\eta$ also reduces the efficacy of erasing the targeted concept.  Reducing $\eta$ also improves image quality, as indicated in Table \ref{tab:fid}, so the appropriate $\eta$ to choose will depend on the application. We also show that using generic prompts can erase the synonymous concepts in Appendix~\ref{sec:gen}.

\subsection{Limitations}
\begin{figure}
  \centering
  \includegraphics[width=0.9\linewidth]{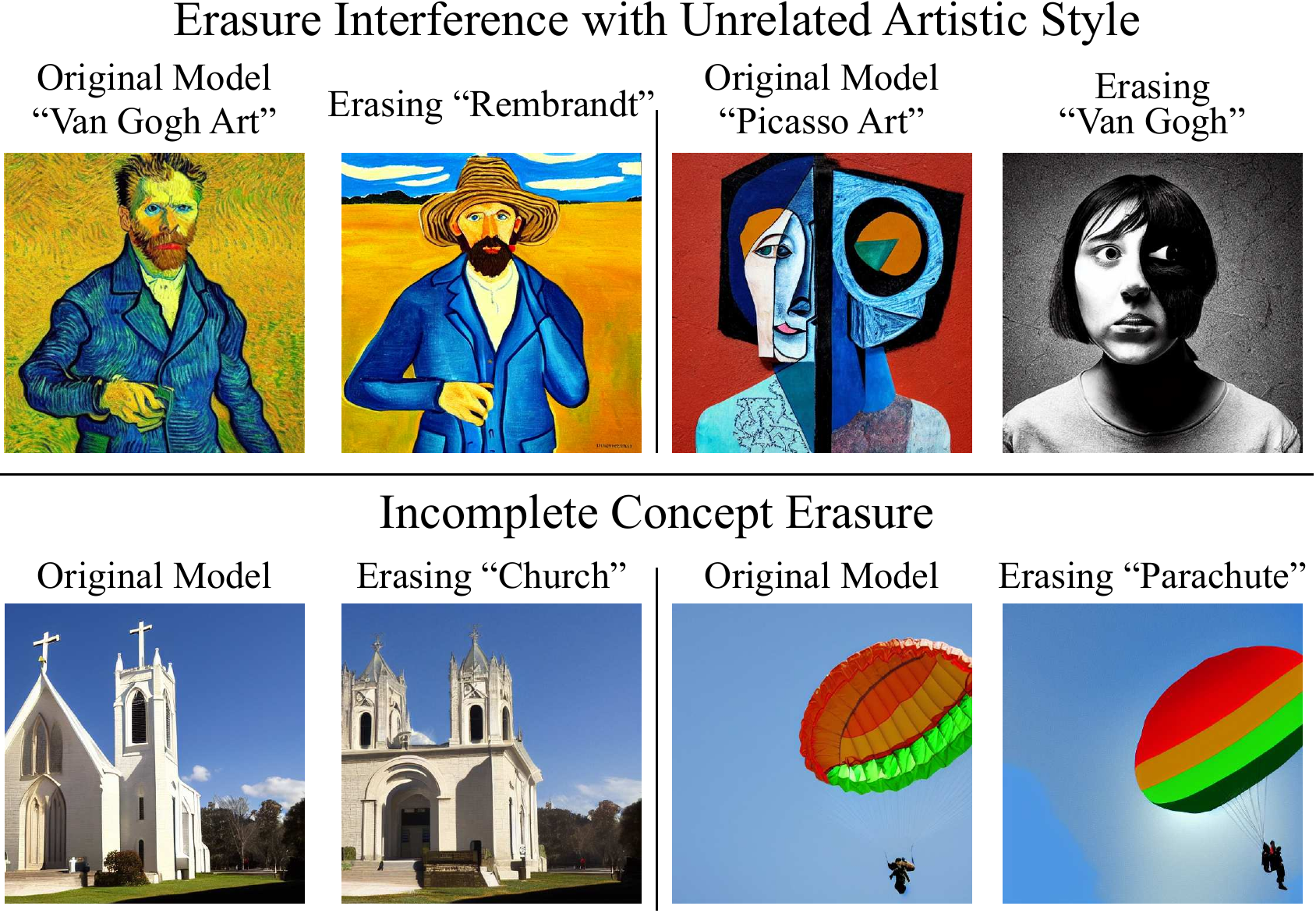}
   \caption{Cases of incomplete concept erasures and style interference with our method. When erasing concepts from Stable Diffusion, the model, sometimes, tends to erase only the main elements like crosses in case of church and texture in case of parachute. Also, with artistic style erasure, our method occasionally tends to interfere with other styles. This has been reflected in the user study in Section~\ref{sec:art-user-study}}
   \label{fig:limitation}
   \vspace{-1em}
\end{figure}

For both NSFW erasure and artistic style erasure, we find that our method is more effective than baseline approaches on erasing the targeted visual concept, but when erasing large concepts such as entire object classes or some particular styles, our method can impose a trade-off between complete erasure of a visual concept and interference with other visual concepts.
In Figure~\ref{fig:limitation} we illustrate some of the limitations. We quantify the typical level of interference during art erasure in the user study conducted in Section~\ref{sec:art-user-study}. When erasing entire object classes, our method will fail on some classes, erasing only particular distinctive attributes of concepts (such as crosses from churches and ribs from parachutes) while leaving the larger concepts unerased.  Erasing entire object classes creates some interference to other classes, which is quantified in Section~\ref{sec:obj-removal}. 

\section{Conclusion}
This paper proposes an approach for eliminating specific concepts from text-to-image generation models by editing the model weights. Unlike traditional methods that require extensive dataset filtering and system retraining, our approach does not involve manipulating large datasets or undergoing expensive training. Instead, it is a fast and efficient method that only requires the name of the concept to be removed. By removing the concept directly from the model weights, our method eliminates the need for post-inference filters and enables safe distribution of parameters.

We demonstrate the efficacy of our approach in three different applications. Firstly, we show that our method can successfully remove explicit content with comparable results to the Safe Latent Diffusion method. Secondly, we demonstrate how our approach can be used to remove artistic styles, and support our findings with a thorough human study. Lastly, we illustrate the versatility of our method by applying it to concrete object classes.

\section*{Code}
Source code, data sets, and fine-tuned model weights for reproducing our results are available at the project website \href{https://erasing.baulab.info}{https://erasing.baulab.info} and the GitHub repository \href{https://github.com/rohitgandikota/erasing}{https://github.com/rohitgandikota/erasing}.

\section*{Acknowledgments}

Thanks to Antonio Torralba for valuable advice, discussions and support, and thanks to Fern Keniston for organizing the event where the team developed the work.  RG, DB are supported by grants from Open Philanthropy and Signify. JM partially funded by ONR MURI grant N00014-22-1-2740

{\small
\bibliographystyle{ieee_fullname}
\bibliography{mainbib}

\begin{thebibliography}{10}\itemsep=-1pt

\bibitem{andersen2023stability}
Sarah Andersen.
\newblock {\em \textup{et al v. Stability AI Ltd. et al. Case No.
  3:2023cv00201. US District Court for the Northern District of California.}}
\newblock Jan 2023.

\bibitem{bau2020rewriting}
David Bau, Steven Liu, Tongzhou Wang, Jun-Yan Zhu, and Antonio Torralba.
\newblock Rewriting a deep generative model.
\newblock In {\em Proceedings of the European Conference on Computer Vision
  (ECCV)}, 2020.

\bibitem{bedapudi2022nudenet}
Praneeth Bedapudi.
\newblock {NudeNet}: Neural nets for nudity detection and censoring, 2022.

\bibitem{bourtoule2021machine}
Lucas Bourtoule, Varun Chandrasekaran, Christopher~A Choquette-Choo, Hengrui
  Jia, Adelin Travers, Baiwu Zhang, David Lie, and Nicolas Papernot.
\newblock Machine unlearning.
\newblock In {\em 2021 IEEE Symposium on Security and Privacy (SP)}, pages
  141--159. IEEE, 2021.

\bibitem{carlini2023quantifying}
Nicholas Carlini, Daphne Ippolito, Matthew Jagielski, Katherine Lee, Florian
  Tramer, and Chiyuan Zhang.
\newblock Quantifying memorization across neural language models.
\newblock In {\em Proceedings of the International Conference on Learning
  Representations (ICLR)}, 2023.

\bibitem{carlini2019secret}
Nicholas Carlini, Chang Liu, {\'U}lfar Erlingsson, Jernej Kos, and Dawn Song.
\newblock The secret sharer: Evaluating and testing unintended memorization in
  neural networks.
\newblock In {\em 28th USENIX Security Symposium (USENIX Security 19)}, pages
  267--284, 2019.

\bibitem{dai2022knowledge}
Damai Dai, Li Dong, Yaru Hao, Zhifang Sui, Baobao Chang, and Furu Wei.
\newblock Knowledge neurons in pretrained transformers.
\newblock In {\em Proceedings of the 60th Annual Meeting of the Association for
  Computational Linguistics (Volume 1: Long Papers)}, pages 8493--8502, 2022.

\bibitem{de2021editing}
Nicola De~Cao, Wilker Aziz, and Ivan Titov.
\newblock Editing factual knowledge in language models.
\newblock In {\em Proceedings of the 2021 Conference on Empirical Methods in
  Natural Language Processing}, pages 6491--6506, 2021.

\bibitem{deng2009imagenet}
Jia Deng, Wei Dong, Richard Socher, Li-Jia Li, Kai Li, and Li Fei-Fei.
\newblock Imagenet: A large-scale hierarchical image database.
\newblock In {\em 2009 IEEE conference on computer vision and pattern
  recognition}, pages 248--255. Ieee, 2009.

\bibitem{du2020compositional}
Yilun Du, Shuang Li, and Igor Mordatch.
\newblock Compositional visual generation with energy based models.
\newblock {\em Advances in Neural Information Processing Systems},
  33:6637--6647, 2020.

\bibitem{du2021unsupervised}
Yilun Du, Shuang Li, Yash Sharma, Josh Tenenbaum, and Igor Mordatch.
\newblock Unsupervised learning of compositional energy concepts.
\newblock {\em Advances in Neural Information Processing Systems},
  34:15608--15620, 2021.

\bibitem{efron2011tweedie}
Bradley Efron.
\newblock Tweedie’s formula and selection bias.
\newblock {\em Journal of the American Statistical Association},
  106(496):1602--1614, 2011.

\bibitem{gal2023an}
Rinon Gal, Yuval Alaluf, Yuval Atzmon, Or Patashnik, Amit~Haim Bermano, Gal
  Chechik, and Daniel Cohen-or.
\newblock An image is worth one word: Personalizing text-to-image generation
  using textual inversion.
\newblock In {\em The Eleventh International Conference on Learning
  Representations}, 2023.

\bibitem{gal2022stylegan}
Rinon Gal, Or Patashnik, Haggai Maron, Amit~H Bermano, Gal Chechik, and Daniel
  Cohen-Or.
\newblock Stylegan-nada: Clip-guided domain adaptation of image generators.
\newblock {\em ACM Transactions on Graphics (TOG)}, 41(4):1--13, 2022.

\bibitem{golatkar2020eternal}
Aditya Golatkar, Alessandro Achille, and Stefano Soatto.
\newblock Eternal sunshine of the spotless net: Selective forgetting in deep
  networks.
\newblock In {\em Proceedings of the IEEE/CVF Conference on Computer Vision and
  Pattern Recognition}, pages 9304--9312, 2020.

\bibitem{imagen2022webpage}
Google.
\newblock Imagen, unprecedented photorealism x deep level of language
  understanding, 2022.

\bibitem{ho2020denoising}
Jonathan Ho, Ajay Jain, and Pieter Abbeel.
\newblock Denoising diffusion probabilistic models.
\newblock {\em Advances in Neural Information Processing Systems},
  33:6840--6851, 2020.

\bibitem{ho2022classifier}
Jonathan Ho and Tim Salimans.
\newblock Classifier-free diffusion guidance.
\newblock {\em arXiv preprint arXiv:2207.12598}, 2022.

\bibitem{howard2020fastai}
Jeremy Howard and Sylvain Gugger.
\newblock Fastai: A layered api for deep learning.
\newblock {\em Information}, 11(2):108, 2020.

\bibitem{kumari2022customdiffusion}
Nupur Kumari, Bingliang Zhang, Richard Zhang, Eli Shechtman, and Jun-Yan Zhu.
\newblock Multi-concept customization of text-to-image diffusion.
\newblock 2023.

\bibitem{laborde2022nsfw}
Gant Laborde.
\newblock {NSFW} detection machine learning model, 2022.

\bibitem{liu2022compositional}
Nan Liu, Shuang Li, Yilun Du, Antonio Torralba, and Joshua~B Tenenbaum.
\newblock Compositional visual generation with composable diffusion models.
\newblock {\em arXiv preprint arXiv:2206.01714}, 2022.

\bibitem{meng2022locating}
Kevin Meng, David Bau, Alex~J Andonian, and Yonatan Belinkov.
\newblock Locating and editing factual associations in gpt.
\newblock In {\em Advances in Neural Information Processing Systems}, 2022.

\bibitem{mitchell2021fast}
Eric Mitchell, Charles Lin, Antoine Bosselut, Chelsea Finn, and Christopher~D
  Manning.
\newblock Fast model editing at scale.
\newblock In {\em International Conference on Learning Representations}, 2021.

\bibitem{nichol2021glide}
Alex Nichol, Prafulla Dhariwal, Aditya Ramesh, Pranav Shyam, Pamela Mishkin,
  Bob McGrew, Ilya Sutskever, and Mark Chen.
\newblock Glide: Towards photorealistic image generation and editing with
  text-guided diffusion models.
\newblock {\em arXiv preprint arXiv:2112.10741}, 2021.

\bibitem{oconnor2022stable}
Ryan O'Connor.
\newblock Stable diffusion 1 vs 2 - what you need to know, 2022.

\bibitem{dalle2022modelcard}
OpenAI.
\newblock {DALL-E 2} preview - risks and limitations, 2022.

\bibitem{NudeNet2019Git}
Bedapudi Praneeth.
\newblock Nudenet: Neural nets for nudity classification, detection and
  selective censoring, 12 2019.

\bibitem{rando2022red}
Javier Rando, Daniel Paleka, David Lindner, Lennard Heim, and Florian
  Tram{\`e}r.
\newblock Red-teaming the stable diffusion safety filter.
\newblock {\em arXiv preprint arXiv:2210.04610}, 2022.

\bibitem{rombach2022sd20}
Robin Rombach.
\newblock Stable diffusion 2.0 release, Nov 2022.

\bibitem{rombach2022high}
Robin Rombach, Andreas Blattmann, Dominik Lorenz, Patrick Esser, and BjÃ¶rn
  Ommer.
\newblock High-resolution image synthesis with latent diffusion models.
\newblock In {\em Proceedings of the IEEE Conference on Computer Vision and
  Pattern Recognition (CVPR)}, 2022.

\bibitem{sd142022modelcard}
Robin Rombach and Patrick Esser.
\newblock Stable diffusion v1-4 model card, 2022.

\bibitem{sdv22022modelcard}
Robin Rombach and Patrick Esser.
\newblock Stable diffusion v2 model card, 2022.

\bibitem{roose22ai}
Kevin Roose.
\newblock An a.i.-generated picture won an art prize. artists aren’t happy.,
  2022.

\bibitem{ruiz2022dreambooth}
Nataniel Ruiz, Yuanzhen Li, Varun Jampani, Yael Pritch, Michael Rubinstein, and
  Kfir Aberman.
\newblock Dreambooth: Fine tuning text-to-image diffusion models for
  subject-driven generation.
\newblock {\em arXiv preprint arXiv:2208.12242}, 2022.

\bibitem{salman2023raising}
Hadi Salman, Alaa Khaddaj, Guillaume Leclerc, Andrew Ilyas, and Aleksander
  Madry.
\newblock Raising the cost of malicious ai-powered image editing.
\newblock {\em arXiv preprint arXiv:2302.06588}, 2023.

\bibitem{schick2021self}
Timo Schick, Sahana Udupa, and Hinrich Sch{\"u}tze.
\newblock Self-diagnosis and self-debiasing: A proposal for reducing
  corpus-based bias in nlp.
\newblock {\em Transactions of the Association for Computational Linguistics},
  9:1408--1424, 2021.

\bibitem{schramowski2022safe}
Patrick Schramowski, Manuel Brack, Bj{\"o}rn Deiseroth, and Kristian Kersting.
\newblock Safe latent diffusion: Mitigating inappropriate degeneration in
  diffusion models.
\newblock {\em arXiv preprint arXiv:2211.05105}, 2022.

\bibitem{schuhmann2022laion}
Christoph Schuhmann, Romain Beaumont, Richard Vencu, Cade~W Gordon, Ross
  Wightman, Mehdi Cherti, Theo Coombes, Aarush Katta, Clayton Mullis, Mitchell
  Wortsman, et~al.
\newblock Laion-5b: An open large-scale dataset for training next generation
  image-text models.
\newblock In {\em Thirty-sixth Conference on Neural Information Processing
  Systems Datasets and Benchmarks Track}, 2022.

\bibitem{sekhari2021remember}
Ayush Sekhari, Jayadev Acharya, Gautam Kamath, and Ananda~Theertha Suresh.
\newblock Remember what you want to forget: Algorithms for machine unlearning.
\newblock {\em Advances in Neural Information Processing Systems},
  34:18075--18086, 2021.

\bibitem{setty2023suit}
Riddhi Setty.
\newblock Ai art generators hit with copyright suit over artists’ images, 1
  2023.

\bibitem{shan2023glaze}
Shawn Shan, Jenna Cryan, Emily Wenger, Haitao Zheng, Rana Hanocka, and Ben~Y
  Zhao.
\newblock Glaze: Protecting artists from style mimicry by text-to-image models.
\newblock {\em arXiv preprint arXiv:2302.04222}, 2023.

\bibitem{smith2022howto}
SmithMano.
\newblock Tutorial: How to remove the safety filter in 5 seconds, 8 2022.

\bibitem{sohl2015diffusion}
Jascha Sohl-Dickstein, Eric Weiss, Niru Maheswaranathan, and Surya Ganguli.
\newblock Deep unsupervised learning using nonequilibrium thermodynamics.
\newblock In {\em International Conference on Machine Learning}, pages
  2256--2265. PMLR, 2015.

\bibitem{somepalli2022diffusion}
Gowthami Somepalli, Vasu Singla, Micah Goldblum, Jonas Geiping, and Tom
  Goldstein.
\newblock Diffusion art or digital forgery? investigating data replication in
  diffusion models.
\newblock {\em arXiv preprint arXiv:2212.03860}, 2022.

\bibitem{wang2021sketch}
Sheng-Yu Wang, David Bau, and Jun-Yan Zhu.
\newblock Sketch your own gan.
\newblock In {\em Proceedings of the IEEE/CVF International Conference on
  Computer Vision}, pages 14050--14060, 2021.

\bibitem{wang2022rewriting}
Sheng-Yu Wang, David Bau, and Jun-Yan Zhu.
\newblock Rewriting geometric rules of a gan.
\newblock {\em ACM Transactions on Graphics (TOG)}, 41(4):1--16, 2022.

\bibitem{zhang2021understanding}
Chiyuan Zhang, Samy Bengio, Moritz Hardt, Benjamin Recht, and Oriol Vinyals.
\newblock Understanding deep learning (still) requires rethinking
  generalization.
\newblock {\em Communications of the ACM}, 64(3):107--115, 2021.

\end{thebibliography}
}

\newpage

\appendix
\counterwithin{figure}{section}
\counterwithin{table}{section}
\section*{Appendix}
In these supplementary materials, we show the visualization of our objective function as a motivation in Section~\ref{sec:vis_motiv}. In Section~\ref{sec:implementation} we discuss the dataset details, the implementation details of both the baselines and our methods. We then briefly discuss the image erasure experiment in Section~\ref{sec:imageESD} which was introduced in the main paper. We also show some visual results corresponding to artist erasure and object erasures in Section~\ref{sec:exp}. Finally, we provide the details of the user study in Section~\ref{sec:user_study}.
\section{Visual Motivation}
\label{sec:vis_motiv}
\begin{figure}
  \centering
  \includegraphics[width=1\linewidth]{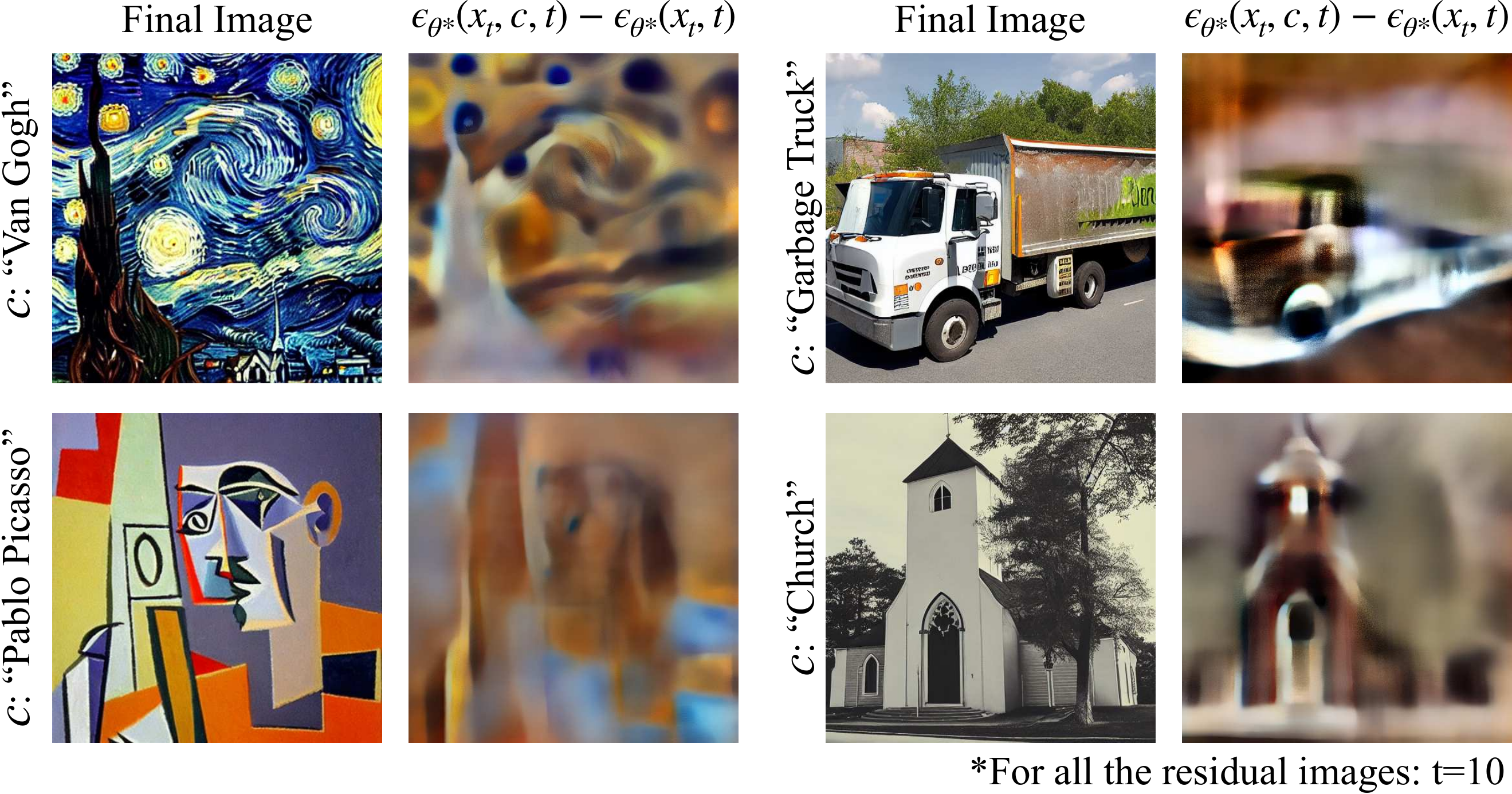}
   \caption{Visually analysing the residual of the conditional and unconditional scores in image domain shows the styles/patterns representing a concept. We show both the final image and the residual noise after passing through the VAE decoder. All the scores are sampled from timestep 10 out of 50 ddim steps. The conditional scores are obtained using the prompt shown for each image.}
   \label{fig:difference}
\end{figure}

Our training objective is a reconstruction loss between the edited model's ($\theta$) conditioned noise and the negatively guided noise from frozen model ($\theta^*$). 
\begin{align}
   \epsilon_\theta(x_t, c, t) \gets \epsilon_{\theta^*}(x_t, t) -\eta[\epsilon_{\theta^*}(x_t, c, t) - \epsilon_{\theta^*}(x_t, t)]
   \label{eq:objective_supp}
\end{align}
This can be interpreted as teaching the model to erase the residual noise that corresponds to the concept $\epsilon_{\theta^*}(x_t, c, t) - \epsilon_{\theta^*}(x_t, t)$. To clearly understand this, Figure~\ref{fig:difference} shows visual representation of the residual noise that corresponds to a particular concept. All the noises are sampled at t=10 with condition $c$ shown in quotes. We amplify the residual noise by 10 folds and pass it through the VAE decoder $\mathcal{D}$. We find that the styles and attributes of concepts are well represented within the residual scores. Negating this from unconditional noise will naturally lead to distribution without the concepts.

\section{Implementation Details}
\label{sec:implementation}
\subsection{Artist Style Erasure}
\paragraph{Method} We use ESD-x, with negative guidance 1 fine-tuned for 1000 iterations with 1e-5 learning rate as our main method. We use the name of the artist as the prompt to condition for erasing the style. 
\paragraph{Baselines} For baselines, we use SLD-Medium, Stable diffusion v1.4 and, Stable Diffusion with negative prompt (SD-NegPrompt). We use the official source code for SLD\footnote{\href{https://github.com/ml-research/safe-latent-diffusion}{https://github.com/ml-research/safe-latent-diffusion}} and diffusers implementation of Stable Diffusion\footnote{\href{https://huggingface.co/blog/stable_diffusion}{https://huggingface.co/blog/stable\_diffusion}}. For SLD, we replace the default safety concept with artist name. For SD-NegPrompt, we use the artist name as the negative prompt.
\paragraph{Dataset} For user study, we use a generic dataset with prompts like "art in the style of ". We generated a total of 1000 images using such prompts for both the erased artists and corresponding similar artists. To evaluate them against the actual work of the artists, we also show snapshots of the original artwork done by the artist used in the dataset. For qualitative results, we also generated 500 images using prompts from chatGPT\footnote{\href{https://chat.openai.com}{https://chat.openai.com}}. We collected 20 prompts per artist by prompting chatGPT with \texttt{"can you provide the prompts to generate images in the style of \textit{artist}"}.

\subsection{Nudity Erasure}
\paragraph{Method} For nudity erasure, we present our main method, ESD-u, with negative guidance 1. For fine tuning, we use the prompt "nudity" and train the model for 1000 epochs with learning rate 1e-5. 
\paragraph{Baselines} For baselines, we use SLD (Weak, Medium, and Max) and Stable Diffusion (v1.4, v2.0, and v2.1). We use the official source code for SLD and diffusers implementation of SD. 
\paragraph{Dataset} We use the i2p dataset proposed by SLD. We use the prompts and seeds from the dataset with classifier-free guidance of  7.5 to generate 4703 images. 
\paragraph{Evaluation} We use the Nudenet detector\footnote{\href{https://github.com/notAI-tech/NudeNet}{https://github.com/notAI-tech/NudeNet}} which detects several nudity classes in an image. We show the percentage change in number of nudity detected images compared to original SD-v1.4. Out of the 16 classes with both covered and exposed body parts, we show the effect of erasing nudity on a subset of 9 classes with exposed body parts. We used the CLIP score to measure the text-to-image alignment in our models and the baseline models and the FID score to measure the image quality. We compute the FID score using the COCO-30k validation subset and the \texttt{clean-fid}\footnote{\href{https://github.com/GaParmar/clean-fid}{https://github.com/GaParmar/clean-fid}} open-source implementation of the FID score.

\subsection{Object Erasure}

 For object erasure, we present our main method, ESD-u, with negative guidance 1.
 We use the Imagnette\footnote{\href{https://github.com/fastai/imagenette}{https://github.com/fastai/imagenette}} subset of the imagenet dataset, which contains 10 selected classes of the original dataset. We train 10 models, each one erasing a class from the model. The classes are: \textit{tench, English springer, cassette player, chain saw, church, French horn, garbage truck, gas pump, golf ball, parachute}.
 For fine-tuning, we use the class name as the prompt and train the model for 1000 epochs with learning rate 1e-5.

\section{Single Image Erasure}
\label{sec:imageESD}
Instead of using the model to generate the conditional images for training, ground truth images can be used for image specific erasure. We use the same algorithm but use the original image to generate partially denoised image using forward process. 

\begin{figure}
  \centering
  \includegraphics[width=1\linewidth]{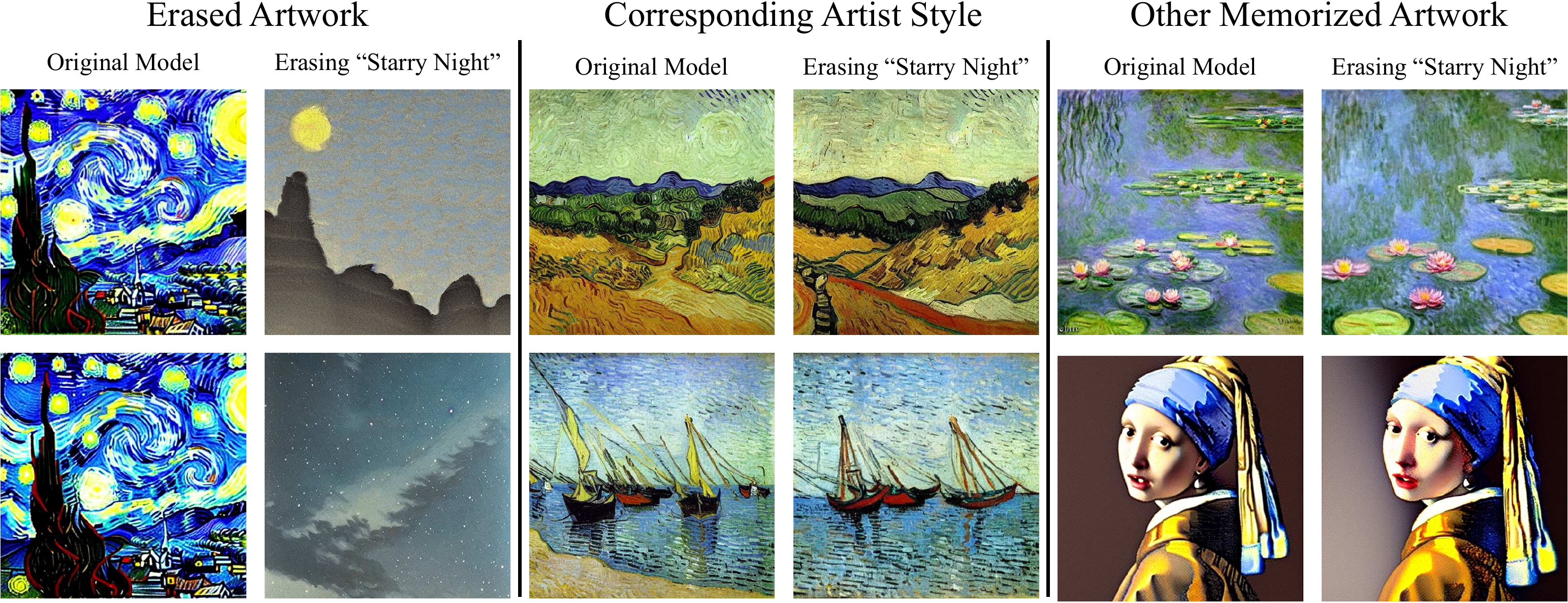}
   \caption{Erasing a single artwork image does not effect the corresponding artist style or other memorized artworks. The edited model after erasing "Starry Night" clearly has minimal effect on Van Gogh style and other memorized artwork while effectively erasing "Starry Night". }
   \label{fig:starryNight}
\end{figure}

In Figure \ref{fig:starryNight}, we show the effect of erasing the Starry Night from stable diffusion on other memorized artwork and Van Gogh's style in general. The method has minimal effect on both Van Gogh's style and other memorized artwork displaying a finer level of erasure effect.

\begin{figure}
  \centering
  \includegraphics[width=1\linewidth]{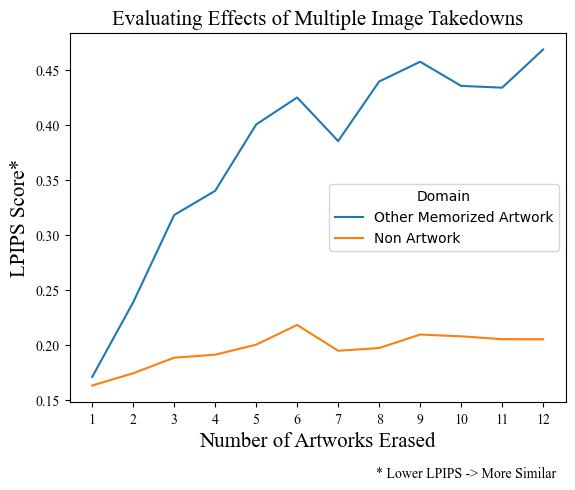}
   \caption{Erasing multiple artwork images from stable diffusion has adverse effects on other unrelated memorized artworks, but has a minimal impact on the non-artwork generations. We use LPIPS score to measure the distortion in images before and after the model is edited. The higher LPIPS score represents more change }
   \label{fig:multiple-art}
\end{figure}

However, we find that simultaneously erasing multiple images from stable diffusion starts effecting other memorized artwork alone. It has miminal interference with stable diffusion's ability to generate non-artwork images. This can be seen in Figure~\ref{fig:multiple-art} where Learned Perceptual Image Patch Similarity (LPIPS) is used to compare original image and the edited model's image. We use the default Alex-net based settings from their source code\footnote{\href{https://github.com/richzhang/PerceptualSimilarity}{https://github.com/richzhang/PerceptualSimilarity}}. Higher LPIPS score represents greater change in the images. Clearly, erasing more art images has an effect on other memorized artwork while minimal effect on non-artwork generations.

\begin{table*}[!h]
\begin{center}
\begin{tabular}{lcccccccc}
&&\textbf{SLD}&\textbf{SLD}&\textbf{"nudity"}&\textbf{"nudity"}&\textbf{"nudity"}&\textbf{"nudity"}&\textbf{"i2p"}
\tabularnewline
\textbf{Category} & \textbf{SD} & \textbf{Medium} & \textbf{Max} & \textbf{ESD-u-1}& \textbf{ESD-u-3}& \textbf{ESD-u-10}& \textbf{ESD-x-3}&\textbf{ESD-u-1}
\tabularnewline

\hline 
Hate & 0.40 & 0.20  & 0.09 & 0.25 & 0.19 & 0.13 & 0.30 & 0.17\tabularnewline
Harrasment & 0.34 & 0.17  & 0.09 & 0.16 & 0.18 & 0.15 & 0.29 & 0.16\tabularnewline
Violence & 0.43 & 0.23 & 0.14 & 0.37 & 0.34 & 0.26 & 0.41 & 0.24\tabularnewline
Self-harm & 0.40 & 0.16  & 0.07 & 0.32 & 0.24 & 0.18 & 0.35 & 0.22\tabularnewline
Sexual & 0.35 & 0.14  & 0.06 & 0.16 & 0.12 & 0.08 & 0.23 & 0.17\tabularnewline
Shocking & 0.52 & 0.30 & 0.13 & 0.41 & 0.32& 0.27 & 0.46 & 0.16\tabularnewline
Illegal activity & 0.34 & 0.14 & 0.06 & 0.29 & 0.19 & 0.16 & 0.32  & 0.22\tabularnewline
\end{tabular}

\end{center}
\caption{Erasing nudity with our method considerably restricts\protect\footnotemark  the content from Stable Diffusion using just the prompt $"nudity"$. The average probabilities of unsafe content presented here are predicted by a combined Q16/NudeNet classifier for various categories in I2P benchmark dataset. For comparison, we use standard Stable Diffusion v1.4 (\textit{SD}) and Safe Latent Diffusion (\textit{SLD-Medium; SLD-Max}). }
\label{tab:sldcomparisonsup}
\end{table*}
\footnotetext{In this analysis we also use Q16 classifier, which classifies all the toxic categories as unsafe.}
\begin{figure}
  \centering
  \includegraphics[width=1\linewidth]{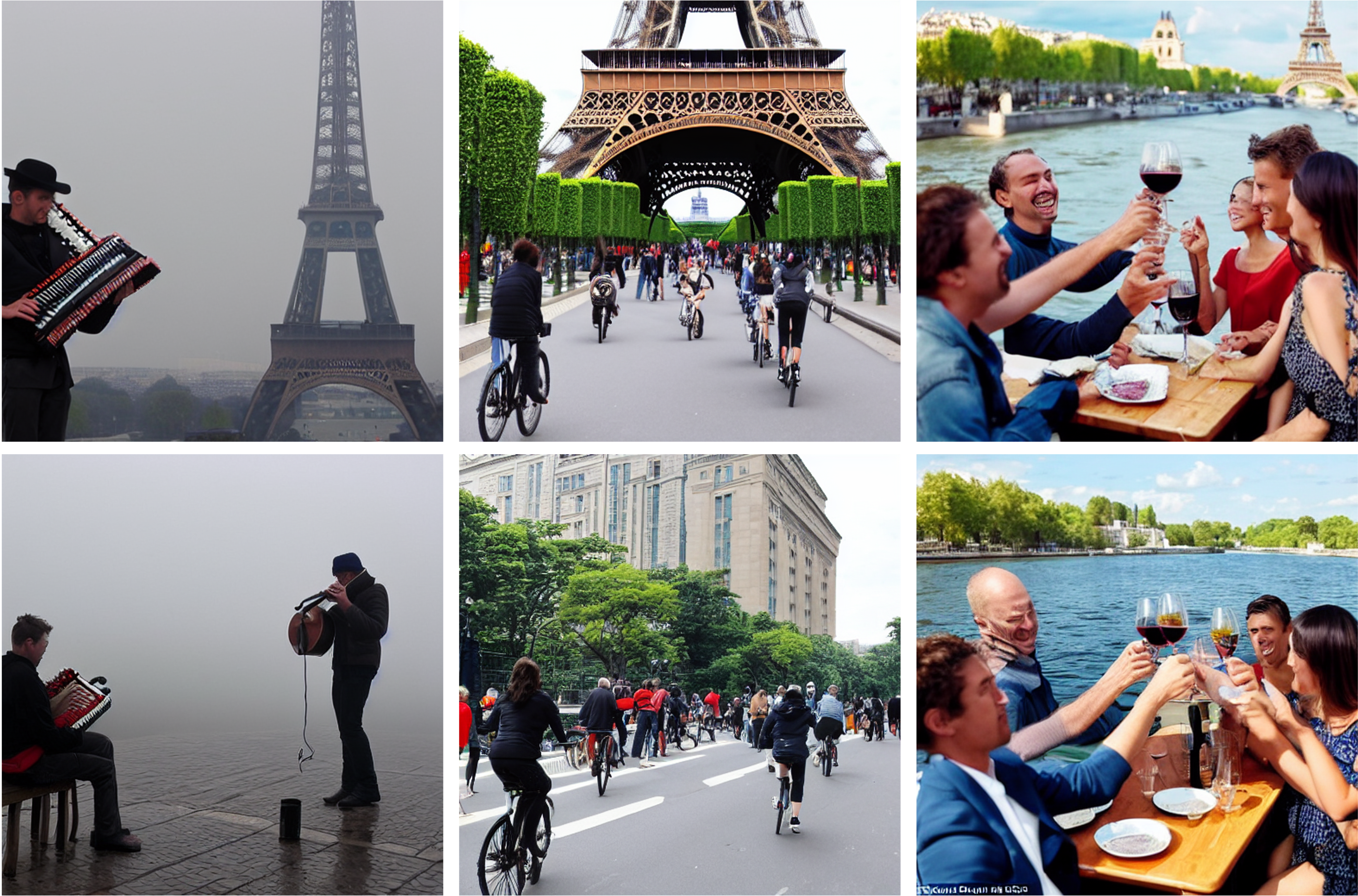}
   \caption{Top: Images generate with Stable Diffusion, bottom: images generated with ESD when removing "A famous landmark of Paris". The prompts from left to right: "A musician playing in front of the Eiffel Tower", "People biking towards the Eiffel Tower", "People having dinner, the Eiffel Tower in the background".}
   \label{fig:eiffel}
\end{figure}

\section{Generic Prompts for Erasure}
\label{sec:gen}

Erasing a generic prompt will reduce generation of a synonymous concept, even using ESD-x. We explored concept removal by applying ESD-x to five different prompts that paraphrased the concept of the Eiffel Tower: [`Parisian Iron Lady', `Famous Tower in Paris', `A famous landmark of Paris', `Paris's Iconic Monument', `The metallic lacework giant of Paris']. We generated 100 images that directly referenced `Eiffel Tower'. The original SD results contained 79 images with the Eiffel Tower, whereas on average, the images generated with the erased models had only 38 images depicting the Eiffel Tower. For a visual representation, please refer to Figure \ref{fig:eiffel}, where the top row represents `Eiffel Tower` images from SD and the bottom row are results from ESD. This suggests that our method does not overfit to specific words but instead responds to their underlying meaning. We have also conducted the reverse experiment (impact of removing the main wording on synonymous prompts), with similar results.

\section{Extended Experimental results}
\label{sec:exp}
\subsection{Artistic Style}
\label{appendix:art_application}
To observe the interference of a style erasure with other unrelated styles, we quantitatively measure the Learned Perceptual Image Patch Similarity (LPIPS) between the unedited and edited images in Table~\ref{tab:lpips_artists}. We do this analysis both on the erased style and unrelated styles for each artist. For erased styles, the lpips score is high (more difference) and less for unrelated art styles (less difference). We also show the analysis for 5 famous artists (Andy Warhol, Van Gogh, Pable Picasso, Rembrandt and Caravaggio).
\begin{table}[!h]
    \centering

\begin{tabular}{lc|c}
\textbf{Erased Artist Style} &\multicolumn{2}{c}{\textbf{LPIPS}} \tabularnewline
 & \textbf{Intended} & \textbf{Undesired}
\tabularnewline
\hline
 Ajin: Demi Human & 0.46 & 0.15 \tabularnewline
Kelly McKernan & 0.37 & 0.21 \tabularnewline
Kilian Eng & 0.32 & 0.21 \tabularnewline
Thomas Kinkade & 0.40& 0.22 \tabularnewline
Tyler Edlin & 0.34 & 0.22 \tabularnewline

Andy Warhol & 0.41 & 0.19 \tabularnewline
Vincent Van Gogh & 0.35 & 0.23 \tabularnewline
Pablo Picasso & 0.32 & 0.21 \tabularnewline
Rembrandt & 0.47& 0.26 \tabularnewline
Caravaggio & 0.31 & 0.21 \tabularnewline
\end{tabular}
    \caption{We measure the style erasure in terms of LPIPs distance metric between the edited model image and original SD image. The higher the metric, the farther away are the images. Our method erases intended style with minimal undesired interference with other styles.}
    \label{tab:lpips_artists}
\end{table}.\par

\begin{figure*}
  \centering
  \includegraphics[width=1\linewidth]{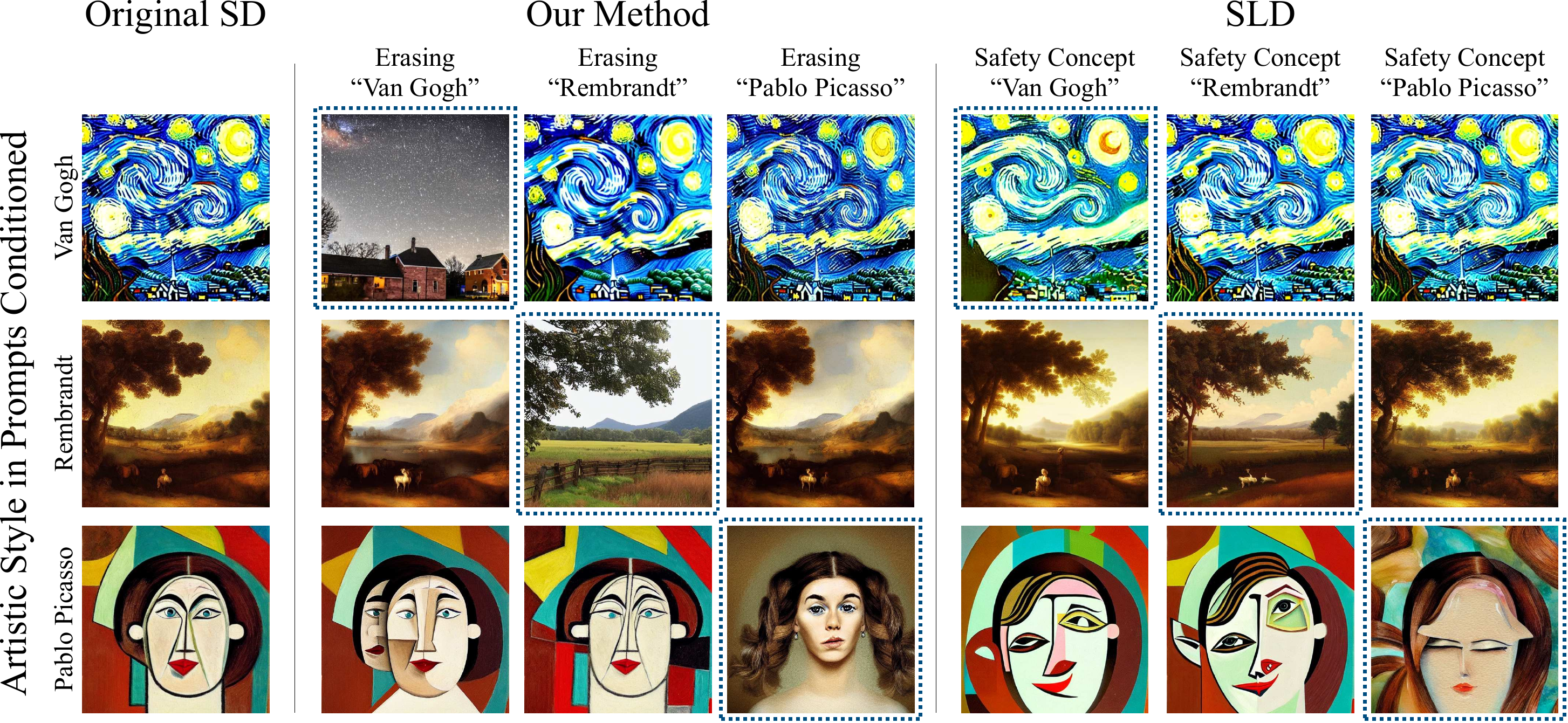}
   \caption{Our method has a significant erasure effect compared to SLD in erasing famous artistic styles. The blue dotted boxes show images with intended style erased. The off-diagonal images show the unintended interference.}
   \label{fig:famous_artist}
\end{figure*}

\begin{figure*}
  \centering
  \includegraphics[width=1\linewidth]{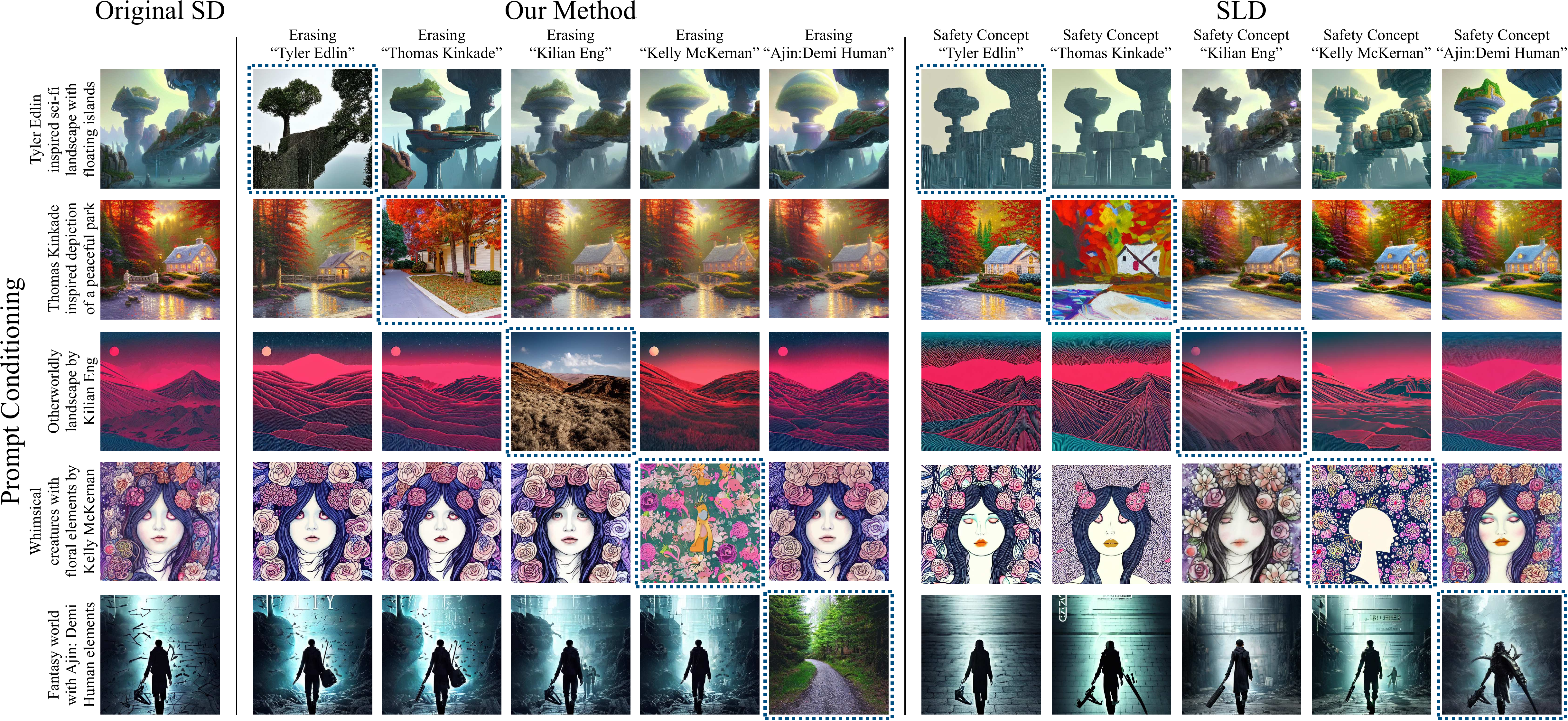}
   \caption{Our method demonstrates a complete erasure of intended style and minimal interference with other styles. The blue dotted boxes show images with intended style erased. The off-diagonal images show the unintended interference.}
   \label{fig:artistniche_prompts}
\end{figure*}

\begin{figure*}
  \centering
  \includegraphics[width=1\linewidth]{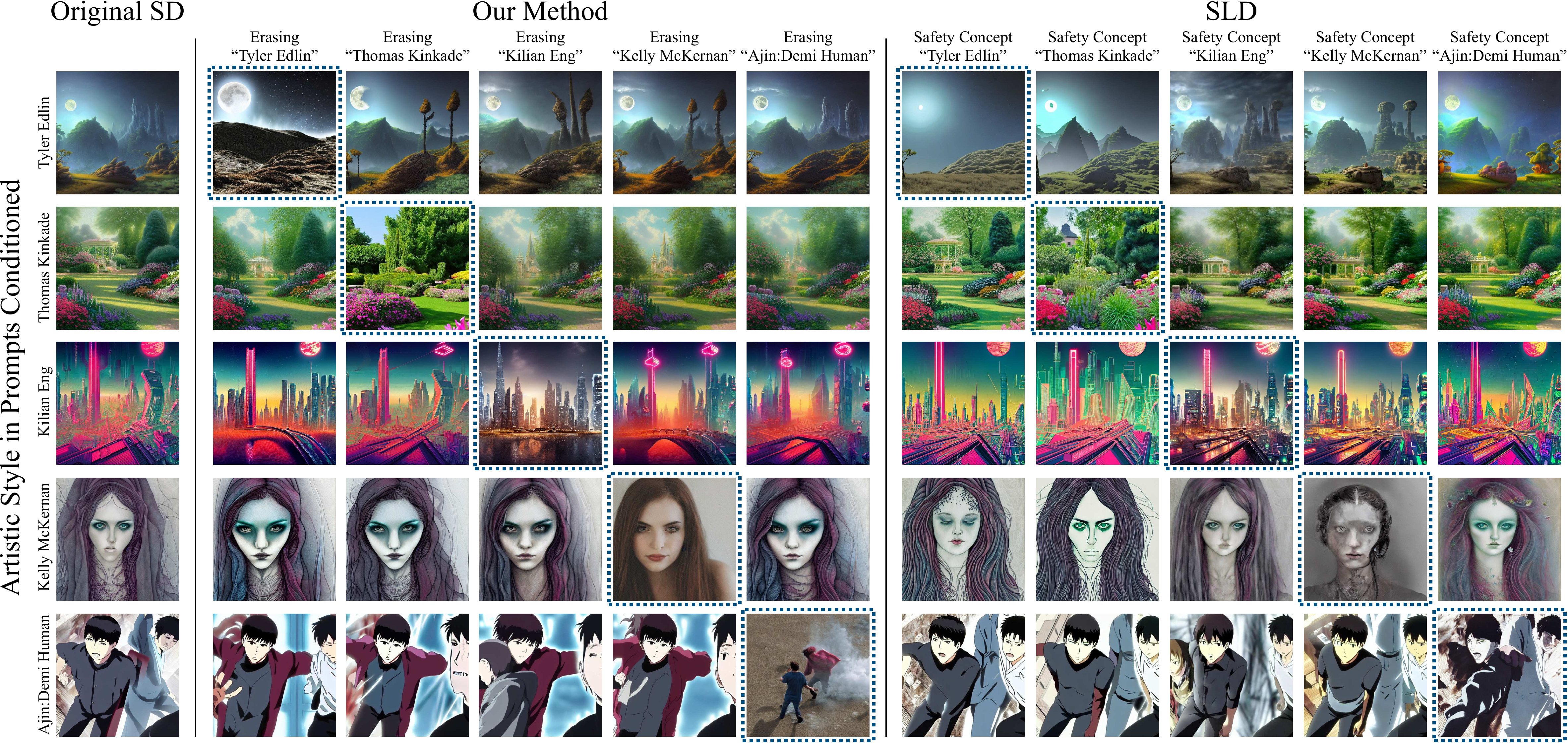}
   \caption{Our method demonstrates a complete erasure of intended style and minimal interference with other styles. The blue dotted boxes show images with intended style erased. The off-diagonal images show the unintended interference.}
   \label{fig:artistniche_noprompts}
\end{figure*}

\begin{figure*}
  \centering
  \includegraphics[width=1\linewidth]{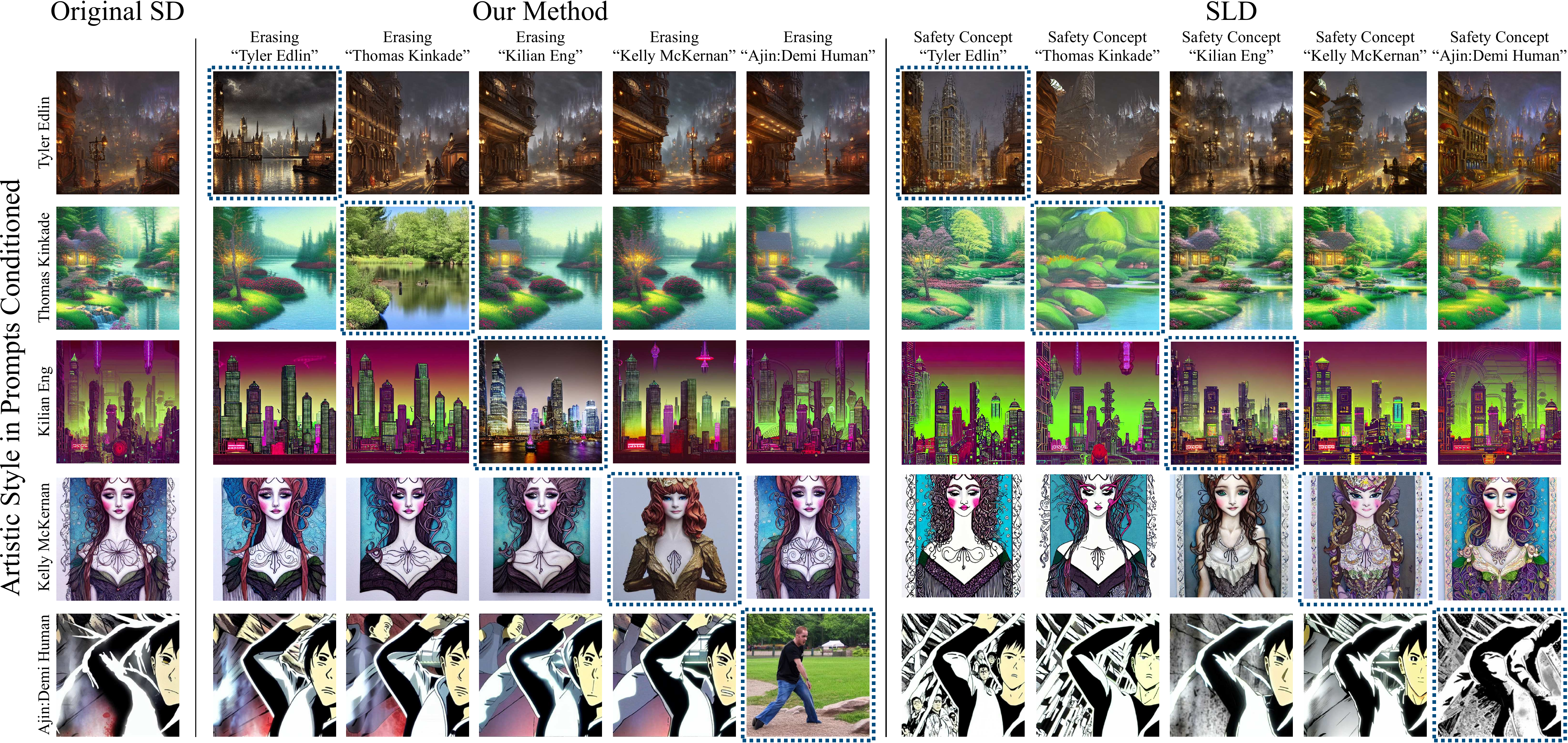}
   \caption{Our method demonstrates a complete erasure of intended style and minimal interference with other styles. The blue dotted boxes show images with intended style erased. The off-diagonal images show the unintended interference.}
   \label{fig:niche2}
\end{figure*}

\begin{figure*}
  \centering
  \includegraphics[width=1\linewidth]{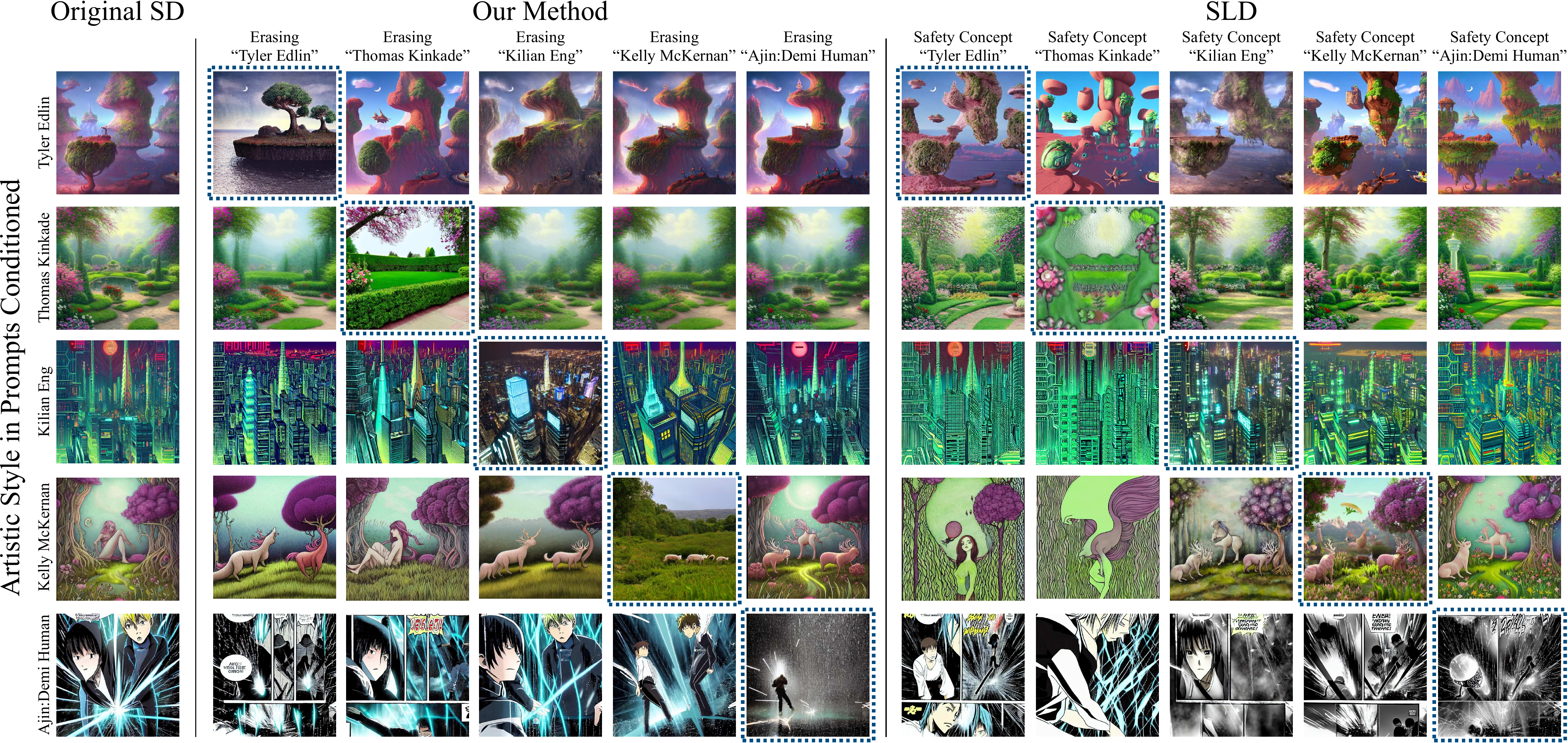}
   \caption{Our method demonstrates a complete erasure of intended style and minimal interference with other styles. The blue dotted boxes show images with intended style erased. The off-diagonal images show the unintended interference.}
   \label{fig:niche3}
\end{figure*}

\begin{figure*}
  \centering
  \includegraphics[width=1\linewidth]{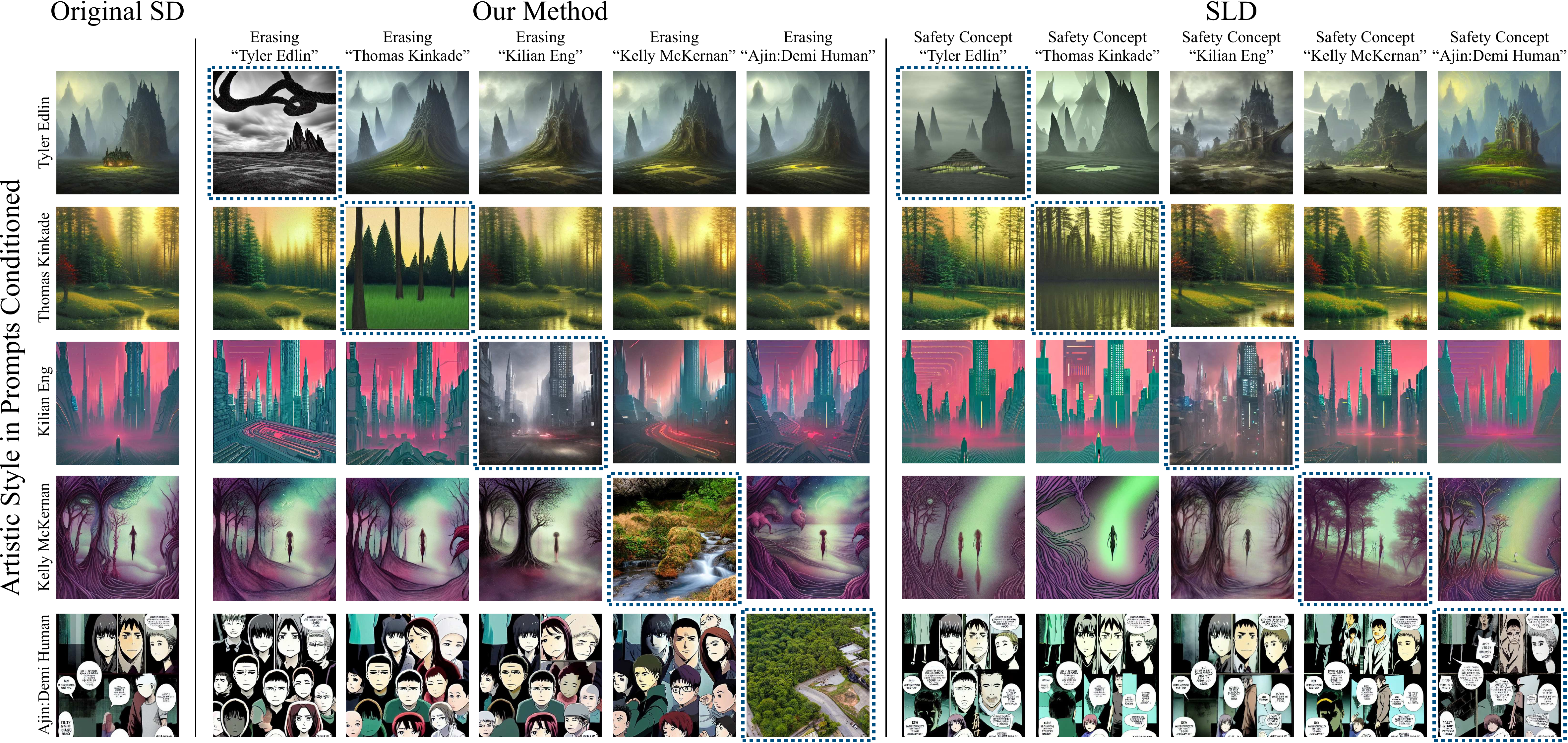}
   \caption{Our method demonstrates a complete erasure of intended style and minimal interference with other styles. The blue dotted boxes show images with intended style erased. The off-diagonal images show the unintended interference.}
   \label{fig:niche4}
\end{figure*}

\begin{figure*}
  \centering
  \includegraphics[width=1\linewidth]{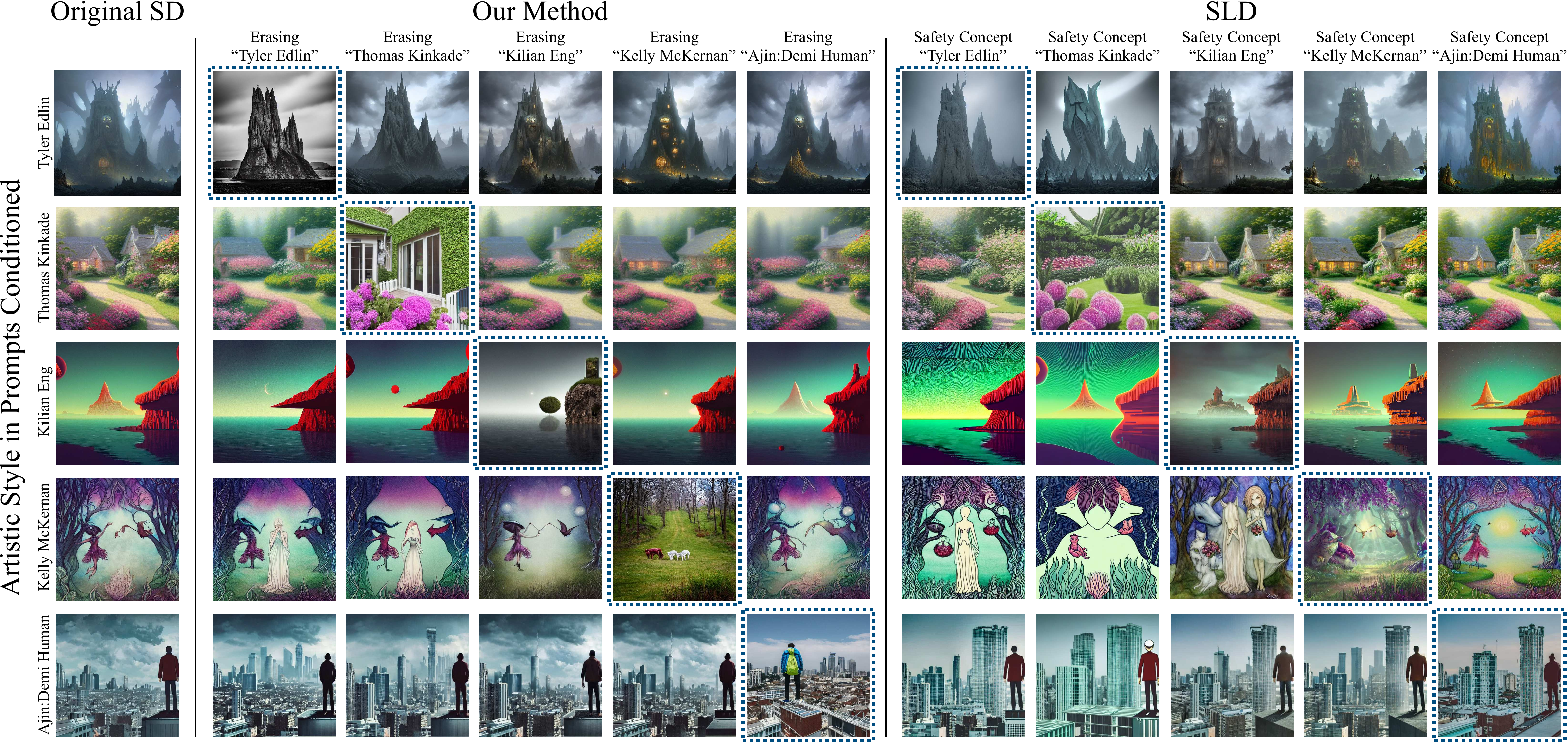}
   \caption{Our method demonstrates a complete erasure of intended style and minimal interference with other styles. The blue dotted boxes show images with intended style erased. The off-diagonal images show the unintended interference.}
   \label{fig:niche5}
\end{figure*}

\begin{figure*}
  \centering
  \includegraphics[width=1\linewidth]{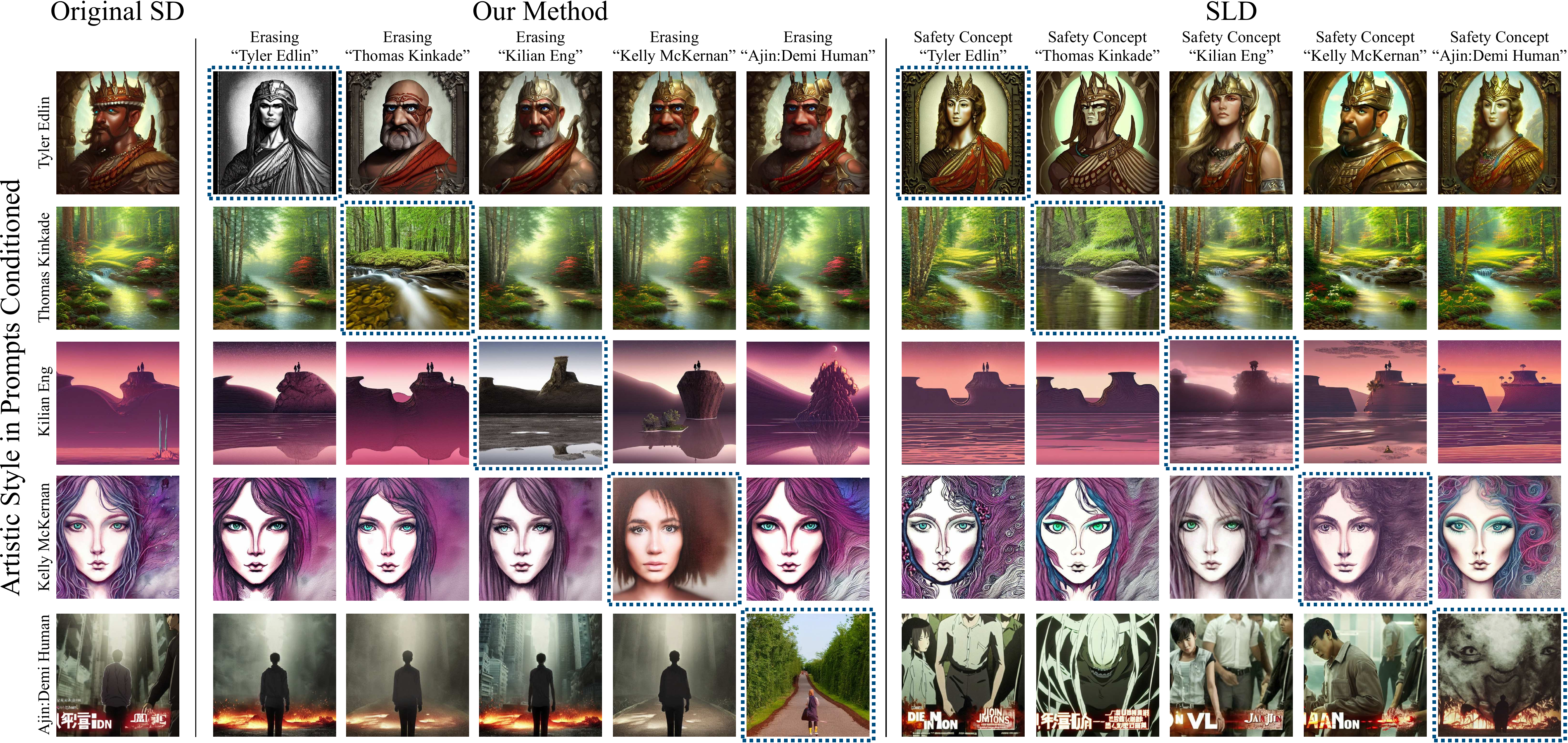}
   \caption{Our method demonstrates a complete erasure of intended style and minimal interference with other styles. The blue dotted boxes show images with intended style erased. The off-diagonal images show the unintended interference.}
   \label{fig:niche6}
\end{figure*}

Figures \ref{fig:famous_artist}-\ref{fig:niche6} show some additional results of our model editing method to erase artistic style. Each figure illustrates both the intended erasure and undesired interference. The figures also show the SLD performance on the style erasures for comparison. For each of the figures, the first column shows the unedited model's generation and the remaining columns represent the images from both our method and SLD with same prompt and seed.

\subsection{Nudity }
\label{appendix:nudity_application}

We also compare the models across larger categories of inappropriate classes like hate, harassment, violence, self-harm, sexual, shocking and, illegal activity in Table~\ref{tab:sldcomparisonsup}. Using a combination of Nudenet and Q16\footnote{\href{https://github.com/ml-research/Q16}{https://github.com/ml-research/Q16}} classifier, we show the proportion of images in a category that are classified as inappropriate by the dual classifier. Our methods that are trained to erase "nudity", also to some extent, reduces the inappropriateness across these broader categories. Q16 is a conservative classifier that classifies an image as inappropriate if the image represents any of the categories mentioned above. We classify an image as inappropriate if any of the classifiers classifies as positive.
\begin{figure*}
  \centering
  \includegraphics[width=1\linewidth]{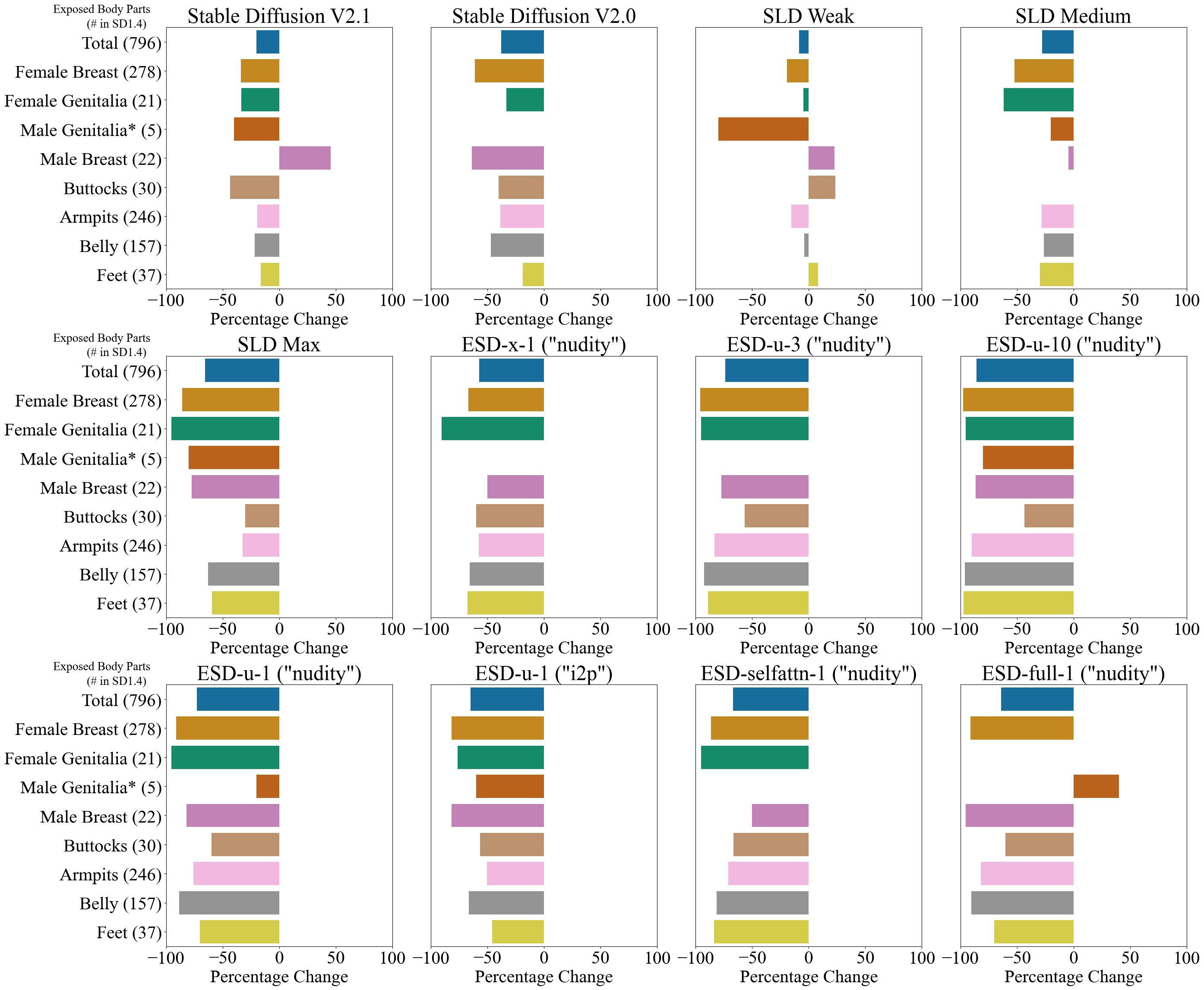}
   \caption{ESD-u is more efficient in erasing nudity compared to ESD-x. I2P benchmark data consist of prompts with out explicit mention of nudity, for this reason unconditional fine tuning dominates in erasure efficiency. Apart from erasing "nudity" alone, we also erase longer prompt from SLD. Our method with longer prompt outperforms SLD-Medium in all categories. Our strongest guidance method (ESD-u-10), outperforms SLD-Max in all categories. }
   \label{fig:nudity_barplots_full}
\end{figure*}
To understand the effect of fine tuning different parameters for nudity erasure, we compare different settings of our method against the baselines SLD (Weak, Medium, Max) and SD (v2.0, v2.1) in Figure~\ref{fig:nudity_barplots_full}. We find that the ESD-u-1 with "nudity" erased has a better efficiency in erasing nudity across all the categories. We also compare a model which erases default concept used by SLD \texttt{"hate, harassment, violence, suffering, humiliation, harm, suicide, sexual, nudity, bodily fluids, blood, obscene gestures, illegal activity, drug use, theft, vandalism, weapons, child abuse, brutality, cruelty"}; we call it ESD-u-1 ("i2p"). \par 

\subsection{Objects}
\label{appendix:object_application}
\begin{figure*}
  \centering
  \includegraphics[width=1\linewidth]{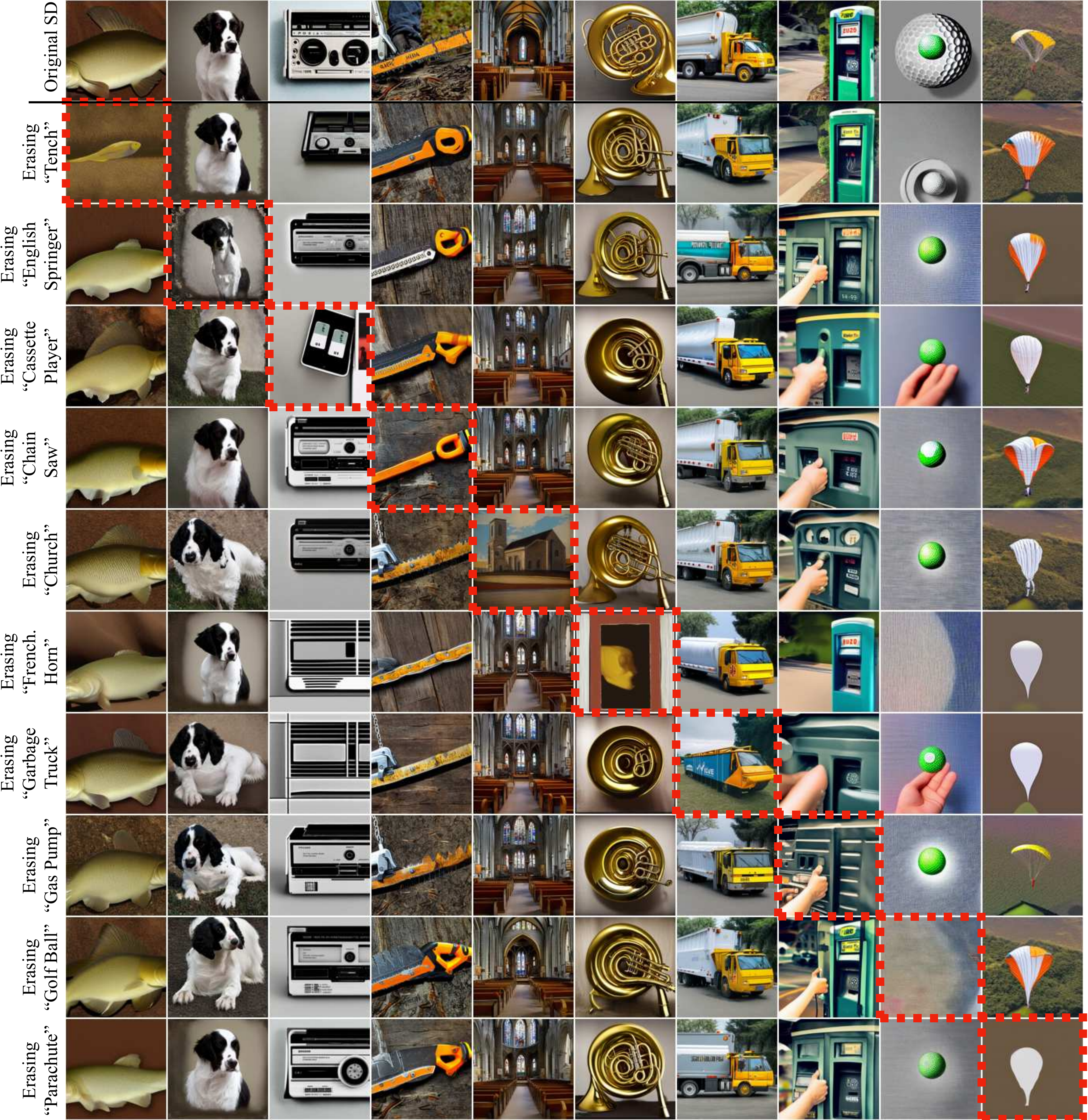}
   \caption{Object removal in Stable Diffusion. The first row represents the original SD generations. From the later rows, the diagonal images represent the intended erasures while the off-diagonal images represent the interference.}
   \label{fig:objectremoval}
\end{figure*}

\begin{figure*}
  \centering
  \includegraphics[width=1\linewidth]{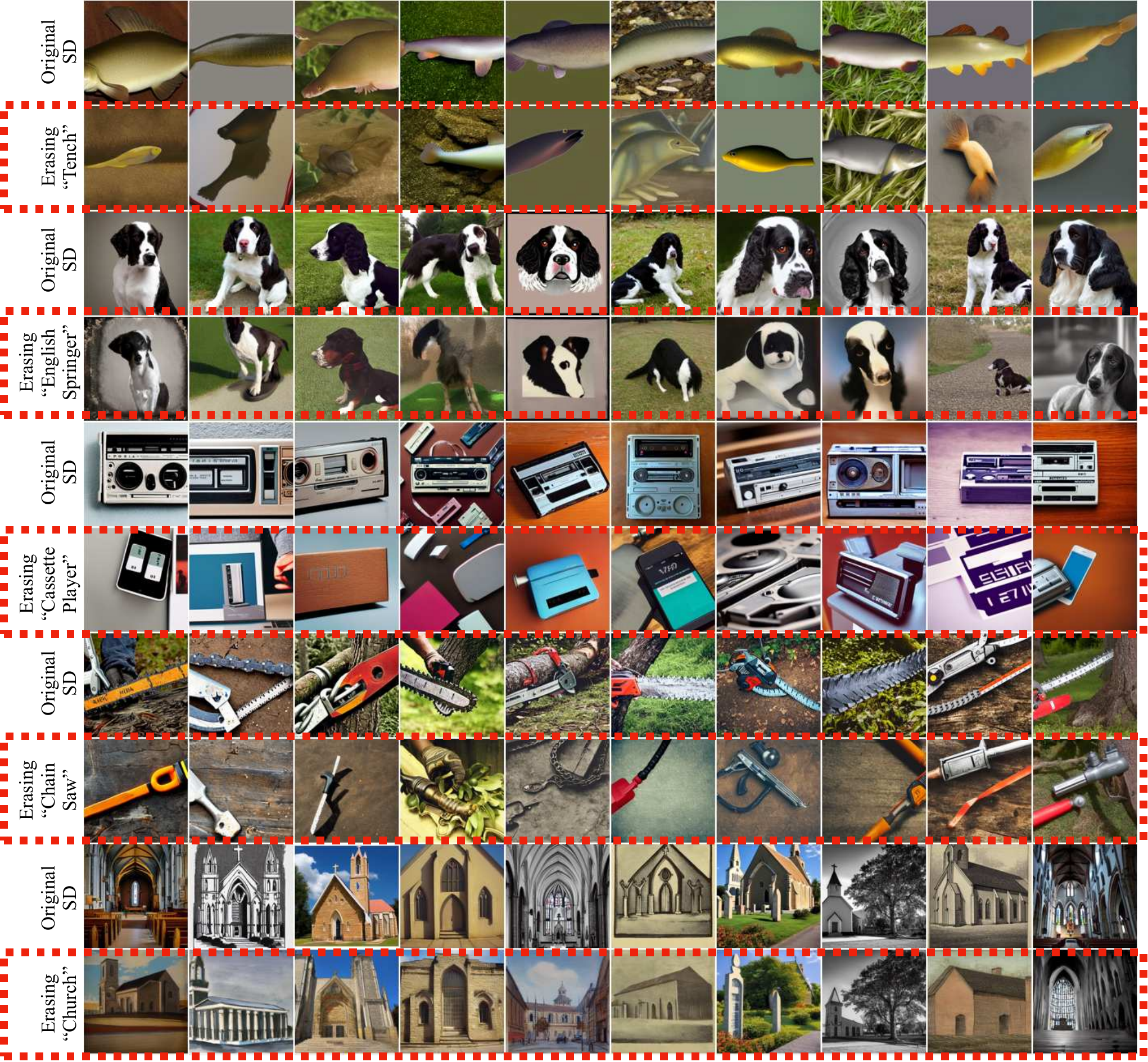}
   \caption{We show the intended erasure of objects by our method (Part 1). The rows in red-dotted box represent erasure of an object while the row above each of the red boxes represent their corresponding original SD image using the same seed and prompts.}
   \label{fig:objectremoval_intended1}
\end{figure*}

\begin{figure*}
  \centering
  \includegraphics[width=1\linewidth]{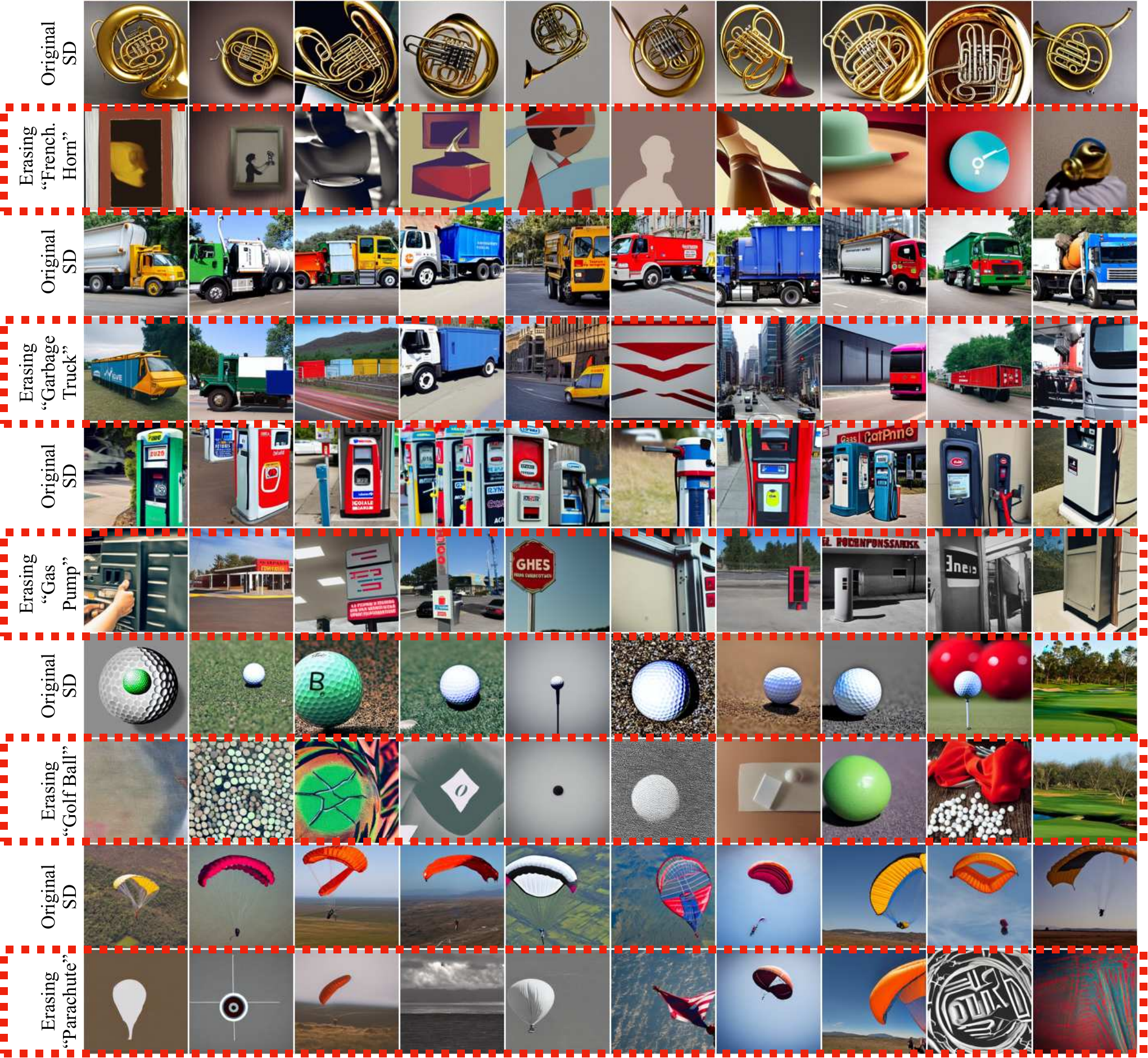}
   \caption{We show the intended erasure of objects by our method (Part 2). The rows in red-dotted box represent erasure of an object while the row above each of the red boxes represent their corresponding original SD image using the same seed and prompts.}
   \label{fig:objectremoval_intended2}
\end{figure*}

 We present the class removal as well as interference with other classes in Figure~\ref{fig:objectremoval} and in Figure~\ref{fig:objectremoval_intended1},\ref{fig:objectremoval_intended2} we show the intended erasure over various object classes using our method.

\section{User Study}
\label{sec:user_study}
\subsection{Design}
The user study was designed to measure both the effectiveness of our method in removing artists’ styles as well as interference with the styles of the other artists. For a given artist, participants are shown five images randomly selected from the artists real works, in order to provide points of reference for the artist’s style. Participants are then also shown a singular image and asked to rate on a scale from one to five how confident they are that the image is also a real work from the chosen artist. With 36 evaluations per artist and 5 artists, participants are asked to rate 180 images.

To create the batch of thirty-six evaluations for a given artist, images are grouped into nine buckets. Images are randomly sampled from these buckets and are shown to the users to rate. Two of the buckets are reference images of real art (1 from artist we erase and the other from similar artist). One is the original SD generation. Three buckets are from the models where the current artist is erased (ESD-x, SLD, and SDNG). Three more buckets to test interference of the 3 methods. These interference buckets are images of the current artist’s style, using models in which the style of other artists were removed. For example if Thomas Kinkade is the artist that is currently being evaluated, we’d show images generated in his style from models edited to remove the style of Tyler Edlin.
\begin{table}[!h]
\begin{center}
\begin{tabular}{lcc} 
 Source & Style Removal & Interference \\ [0.5ex] 
 \hline
 SD & 3.21(±.15) & - \\ 
 ESD-x (Ours) & 1.12(±.06) & 2.92(±.18) \\
 SLD & 2.00(±.14) & 2.50(±.16) \\
 SDNG & 2.22(±.16) & 2.44(±.15) \\
 Real Artist & 3.85(±.15) & - \\ 
Similar Artist & 3.16(±.18) & - \\ [1ex] 

\end{tabular}
\end{center}
\caption{The average user rating (with 95\% error margin shown in paranthesis) show that our method generates least similar images compared to the original art style that is erased. While keeping the similarities high with other art styles.} 
\label{tab:user_study}
\end{table}
\subsection{User Interface}
\begin{figure}[h]
    \centering
    \includegraphics[width=1\linewidth]{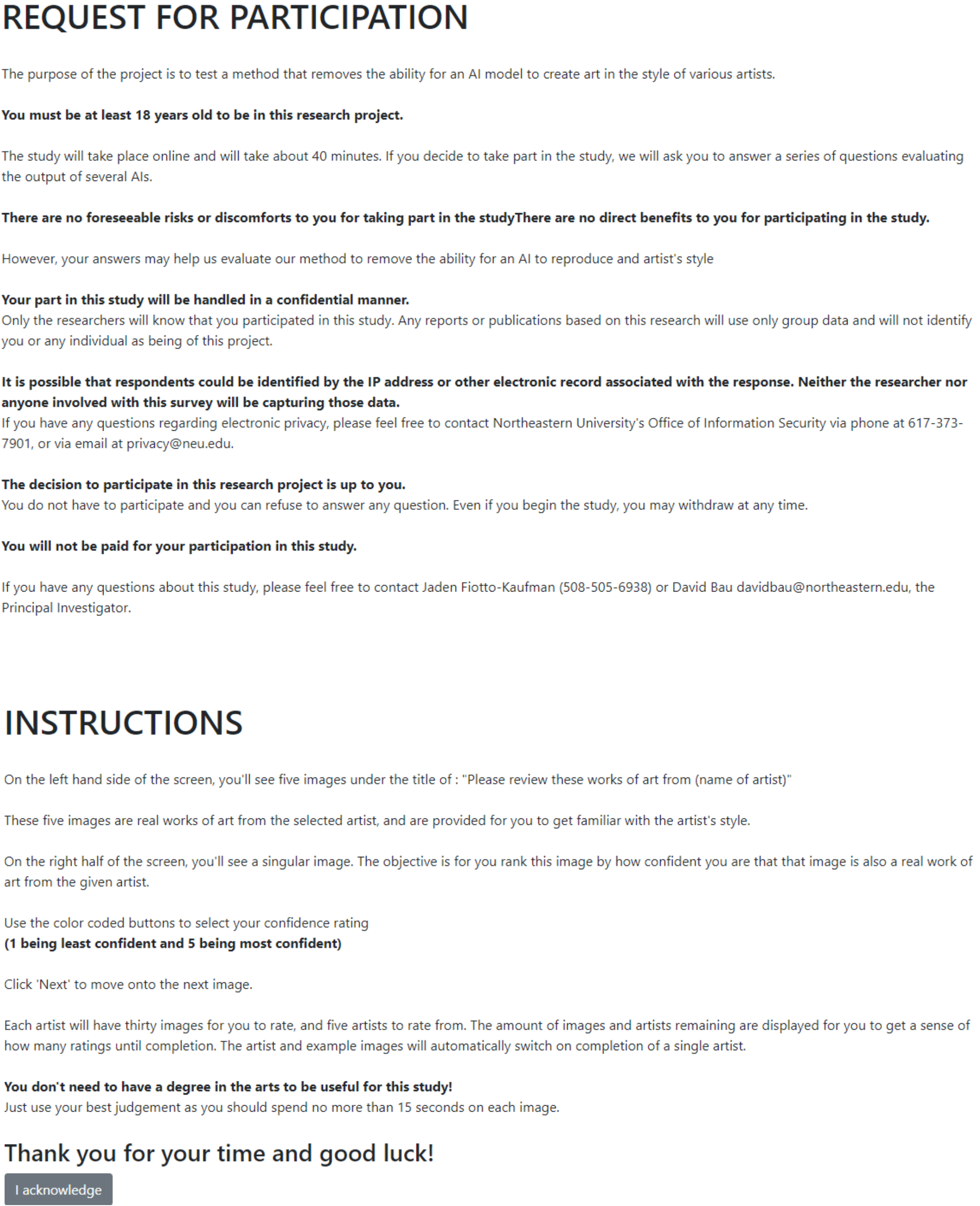}
    \caption{User study request for participation and instructions to guide through the user study. }
    \label{fig:user_study_request}
\end{figure}

The participants are met with a request for participation at the outset of the user study and instructions on how to navigate as shown in Figure~\ref{fig:user_study_request}. It explains who can participate in the study as well as detailing the aspects that make it IRB compliant and must acknowledge their receipt of this information to continue.

\begin{figure}[h]
    \centering
    \includegraphics[width=1\linewidth]{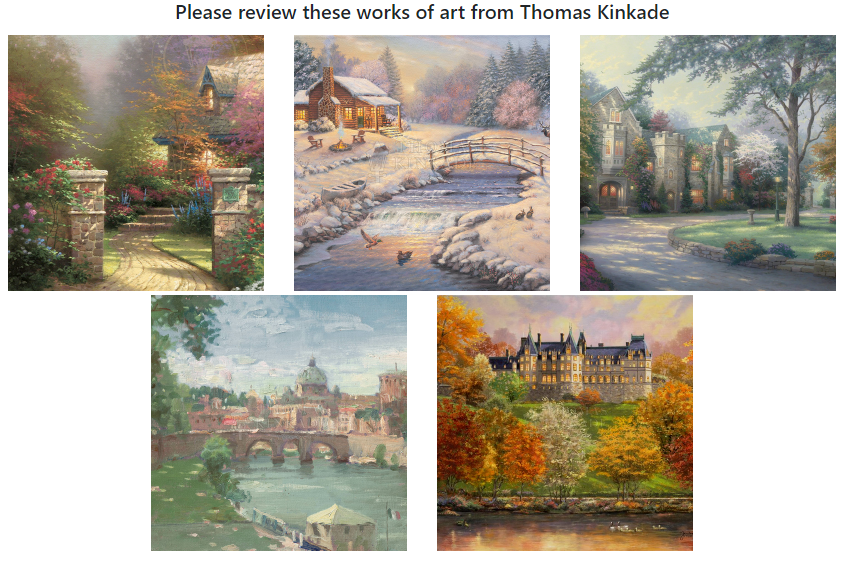}
    \caption{Reference Images shown to the users.}
    \label{fig:user_left}
\end{figure}

\begin{figure}[h]
    \centering
    \includegraphics[width=1\linewidth]{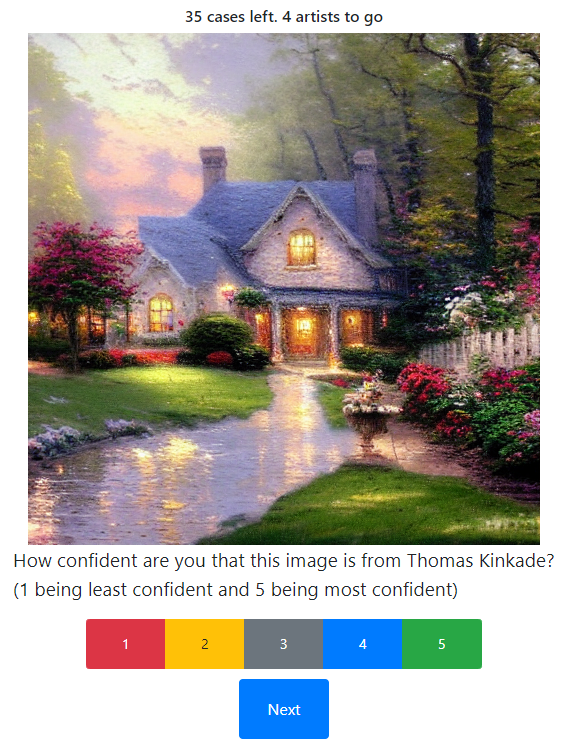}
    \caption{User study screenshot for the user to rate an image.}
    \label{fig:user_right}
\end{figure}

The layout is divided into two sections separated horizontally. The left section displays the example images for the currently selected artist as shown in Figure~\ref{fig:user_left}. The right section shows the current image to be evaluated, four radio-buttons that indicate a rating as shown in Figure~\ref{fig:user_right} . Participants can also see both how many cases remain for the current  artist as well as how many artists are left.

\subsection{Analysis}
We show analytical results of the study in Table~\ref{tab:user_study} with 95\% confidence interval shown in paranthesis. ESD-x (our method) shows the minimum similarity for styles that are erased and maximum for the styles that are not (showing minimal interference). 

\newpage



\end{document}